\documentclass[11pt]{article}
\usepackage{latex/acl}
\usepackage{amsmath}
\usepackage{times}
\usepackage{latexsym}
\usepackage{graphicx}
\usepackage{xcolor}
\usepackage{makecell}
\usepackage{comment}
\usepackage{booktabs}
\usepackage{multirow}
\usepackage{CJKutf8}
\usepackage{graphicx}
\usepackage[T1]{fontenc}
\usepackage{algorithm}
\usepackage{algorithmic}
\usepackage[utf8]{inputenc}
\usepackage{multirow}
\usepackage{enumitem}

\usepackage{microtype}
\usepackage{hyperref}
\usepackage{inconsolata}
\usepackage{amssymb}
\usepackage{amsmath}
\usepackage{booktabs}
\usepackage{lipsum}
\usepackage{colortbl}

\definecolor{firstBest}{rgb}{0.86, 1, 0.86} 

\usepackage{pifont} 

\usepackage{xspace}
\makeatletter
\DeclareRobustCommand\onedot{\futurelet\@let@token\@onedot}
\def\@onedot{\ifx\@let@token.\else.\null\fi\xspace}

\title{ ``See the World, Discover Knowledge'': \\ A Chinese Factuality Evaluation for Large Vision Language Models}

\author{
  \textbf{Jihao Gu}\thanks{Work done during an internship at Alibaba Group.}, 
  \textbf{Yingyao Wang}\thanks{Equal contribution.}, 
  \textbf{Pi Bu}, \\
  \textbf{Chen Wang},
  \textbf{Ziming Wang},
  \textbf{Tengtao Song},
  \textbf{Donglai Wei}, 
  \textbf{Jiale Yuan}, \\
  \textbf{Yingxiu Zhao},
  \textbf{Yancheng He}, 
  \textbf{Shilong Li}, 
  \textbf{Jiaheng Liu},
  \textbf{Meng Cao},
  \textbf{Jun Song}\thanks{Corresponding Author.},  \\
  \textbf{Yingshui Tan},
  \textbf{Xiang Li},
  \textbf{Wenbo Su},
  \textbf{Zhicheng Zheng}
  \textbf{Xiaoyong Zhu},
\textbf{Bo Zheng},\\
  Alibaba Group \\
  {\{gujihao.gjh, wangyingyao.wyy, jsong.sj\}}@taobao.com
}

\begin{document}
\maketitle
\begin{abstract}
The evaluation of factual accuracy in large vision language models (LVLMs) has lagged behind their rapid development, making it challenging to fully reflect these models' knowledge capacity and reliability. In this paper, we introduce the first factuality-based visual question-answering benchmark in Chinese, named \textbf{ChineseSimpleVQA}, aimed at assessing the visual factuality of LVLMs across 8 major topics and 56 subtopics. The key features of this benchmark include a focus on the \textbf{Chinese} language, \textbf{diverse} knowledge types, a \textbf{multi-hop} question construction, \textbf{high-quality} data, \textbf{static} consistency, and \textbf{easy-to-evaluate} through short answers. 
Moreover, we contribute a rigorous data construction pipeline and decouple the visual factuality into two parts: seeing the world (i.e., object recognition) and discovering knowledge. This decoupling allows us to analyze the capability boundaries and execution mechanisms of LVLMs. Subsequently, we evaluate 35 advanced open-source and closed-source models, revealing critical performance gaps within this field. Our evaluation-friendly code \footnote{\url{https://github.com/OpenStellarTeam/ChineseSimpleVQA}} and data \footnote{\url{https://huggingface.co/datasets/OpenStellarTeam/Chinese-SimpleVQA}} have already been open-sourced.
\end{abstract}

\section{Introduction}

\begin{figure}[t]
\begin{center}
\includegraphics[scale=0.4]{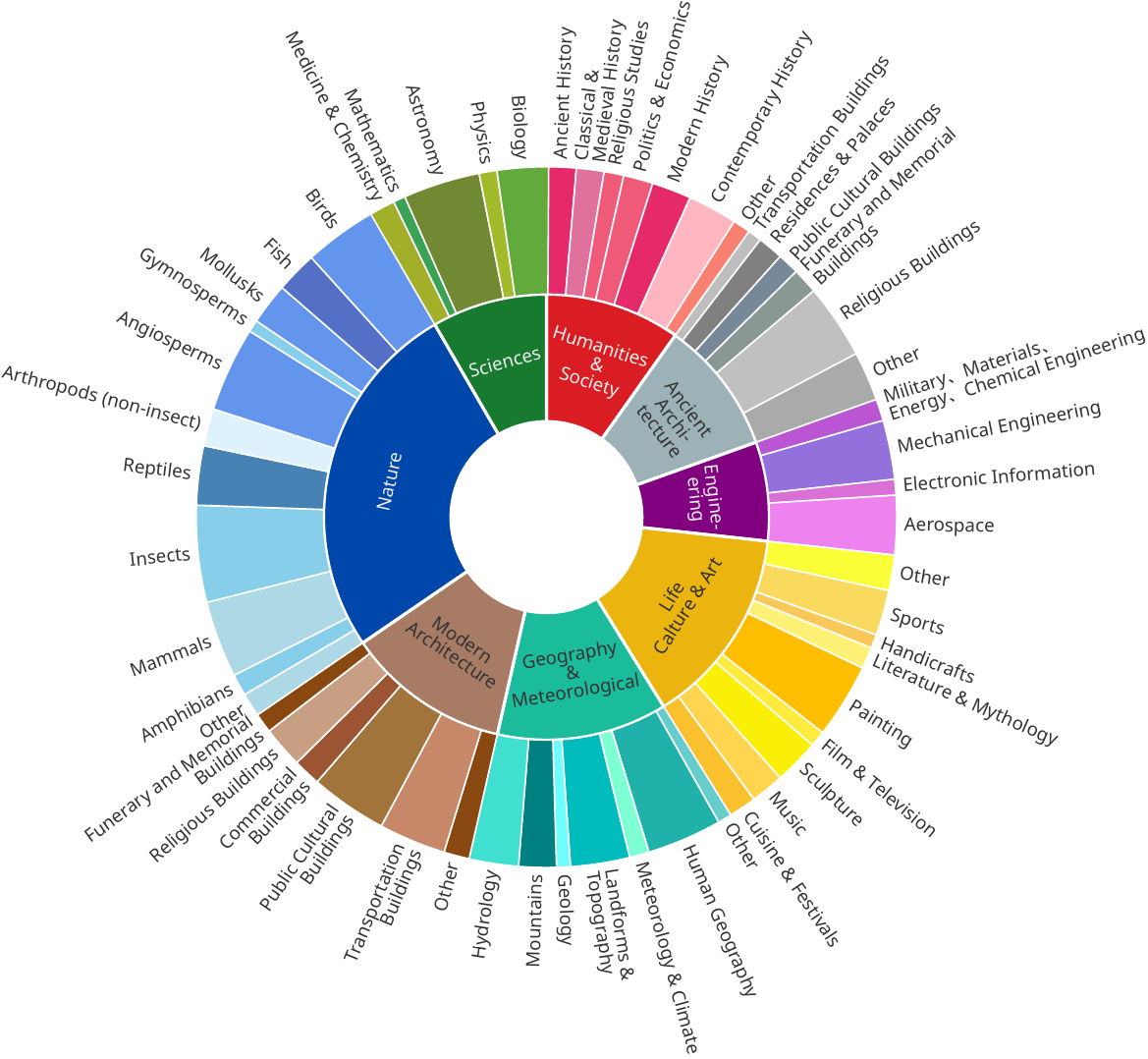}
\caption{The overview of ChineseSimpleVQA, which evaluates visual factuality covering 56 subtopics.}
\label{fig.distribution}
\end{center}
\vspace{-0.6cm}
\end{figure}

\begin{table*}[ht]
\centering
\small 
\resizebox{\textwidth}{!}{
\begin{tabular}{l r l l c c c c c l}
\toprule
\textbf{Benchmark}  & \textbf{Size} & \textbf{Lang} & \textbf{Source} & \textbf{Categories} &\textbf{Vision} & \textbf{Open-ended} & \textbf{Short-form} & \textbf{Metric} \\
\midrule
\begin{tabular}[c]{@{}l@{}}OK-VQA \\ \cite{marino2019ok}  \end{tabular} &  8,062 & en & Exams & 20 & \textcolor{green}{\checkmark} & \textcolor{red}{\texttimes} & \textcolor{red}{\texttimes} & Accuracy \\
\begin{tabular}[c]{@{}l@{}}A-OKVQA \\ \cite{schwenk2022okvqa} \end{tabular} & 125 & zh & Human Collection & 7 & \textcolor{green}{\checkmark} & \textcolor{green}{\checkmark} & \textcolor{red}{\texttimes} & LLM-as-a-Judge \\
\midrule
\begin{tabular}[c]{@{}l@{}}SimpleQA \\ \cite{Wei2024MeasuringSF}  \end{tabular}  & 4,326 & en & Human Writers & 10 &\textcolor{red}{\texttimes} & \textcolor{green}{\checkmark} & \textcolor{green}{\checkmark} & LLM-as-a-Judge \\
\begin{tabular}[c]{@{}l@{}}Chinese SimpleQA \\ \cite{he2024chinese}  \end{tabular}  & 3,000 & zh & \begin{tabular}[c]{@{}l@{}}Self-constructed \\ \& Human Writers  \end{tabular} & 99 &\textcolor{red}{\texttimes} & \textcolor{green}{\checkmark} & \textcolor{green}{\checkmark} & LLM-as-a-Judge \\
\midrule
\begin{tabular}[c]{@{}l@{}}\textbf{ChineseSimpleVQA} \\ \textbf{(Ours)} \end{tabular}  & 2,200 & zh & \begin{tabular}[c]{@{}l@{}}Self-constructed \\ \& Human Writers  \end{tabular} & 56  &  \textcolor{green}{\checkmark} & \textcolor{green}{\checkmark} & \textcolor{green}{\checkmark} & LLM-as-a-Judge \\
\bottomrule
\end{tabular}
}
\caption{Comparisons between our ChineseSimpleVQA and other knowledge-based VQA benchmarks.}
\label{table.1}

\end{table*}

In the multimodal field, ensuring the factual accuracy of responses generated by Large Vision Language Models (LVLMs) presents a significant challenge. Currently, cutting-edge models still produce erroneous outputs that do not align with the image content or provide answers lacking support from knowledge evidence. This issue is known as the ``visual hallucination'', primarily arising from difficulties in multimodal alignment and the lack of visual knowledge. In the realm of Large Language Models (LLMs), the factuality assessment of linguistic hallucination has garnered extensive attention. For instance, OpenAI introduced the SimpleQA benchmark \cite{Wei2024MeasuringSF}, and Alibaba launched a Chinese SimpleQA \cite{he2024chinese}, both of them efficiently evaluate the lengthy responses generated by LLMs containing numerous factual claims through simple questions.

\begin{figure*}[t]
\begin{center}
\includegraphics[scale=0.5]{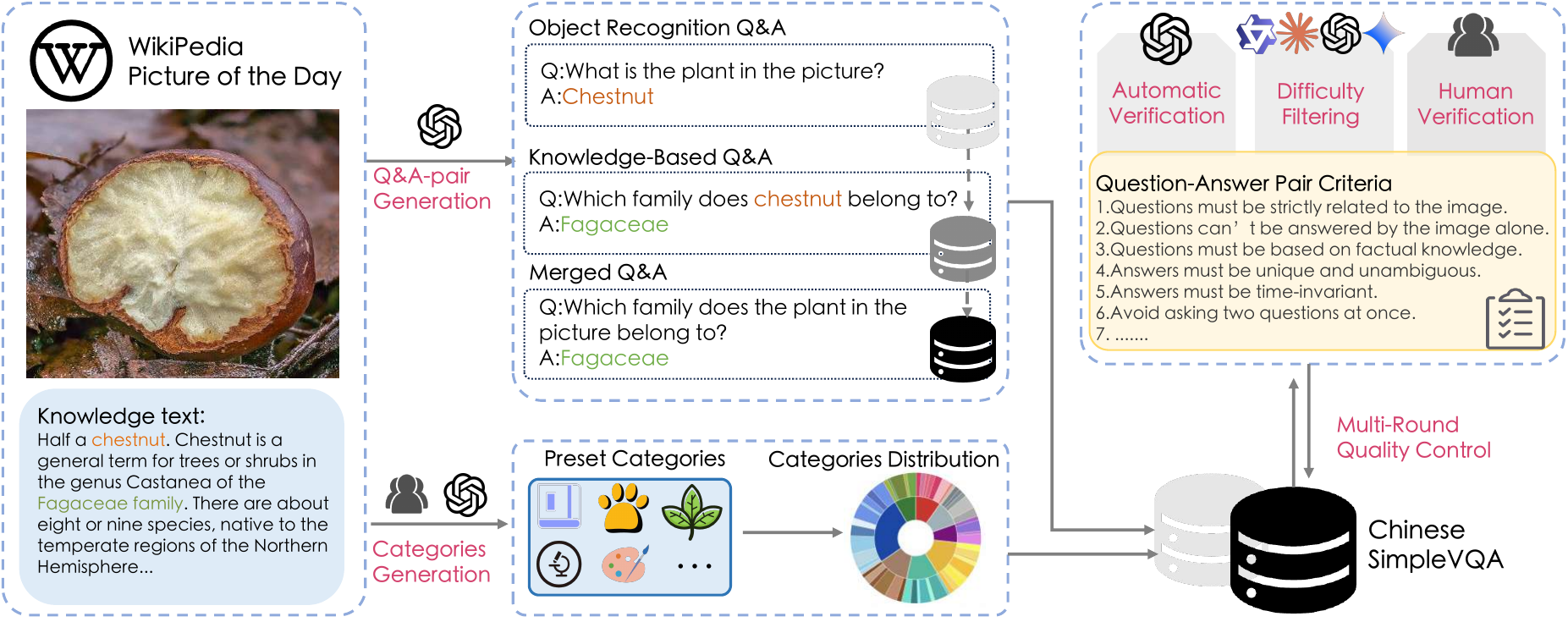}
\caption{The production pipeline of ChineseSimpleVQA consists of automated verification, difficulty filtering, and human verification. This process generates multi-hop questions (i.e. object recognition Q\&A and merged Q\&A).}
\label{fig.method}
\end{center}
\vspace{-0.6cm}
\end{figure*}

However, the factuality assessment concerning visual hallucination has not received adequate attention \cite{fu2024mme}. Researchers have extensively employed benchmarks such as OK-VQA \cite{marino2019ok} and A-OKVQA \cite{schwenk2022okvqa} to assess the factuality capabilities of LVLMs, but both of them have a narrow knowledge coverage (e.g., A-OKVQA only includes 7 categories).
Furthermore, we emphasize that the evaluation of visual factuality should be decoupled into two parts: seeing the world (i.e., object recognition) and discovering knowledge.
As illustrated in Figure \ref{fig.method}, a model must first comprehend that the image depicts a ``chestnut'' to answer the next question, ``What family does it belong to?'' This decoupling of object recognition and knowledge discovery aids in a deeper analysis of the capability boundaries and execution mechanisms of LVLMs.

In this paper, to comprehensively assess the factual knowledge of LVLMs, we present a \textbf{ChineseSimpleVQA} benchmark\footnote{\url{https://chinesesimplevqa.github.io/ChieseSimpleVQA.github.io/}}, which consists of a dataset containing 2,200 high-quality questions across 56 topics, spanning from the humanities to science and engineering, as depicted in Figure \ref{fig.distribution}. Specifically, the key distinguishing features of our proposed ChineseSimpleVQA are as follows:

\begin{itemize}[leftmargin=*]
\item \textbf{Multi-hop:} Visual factuality inquiries are decomposed into two steps: object recognition and knowledge assessment. This multi-hop strategy allows us to analyze the capability boundaries and execution mechanisms of LVLMs.
\item \textbf{Diverse:} ChineseSimpleVQA emphasizes the Chinese language and covers 8 major topics (i.e., ``Nature'', ``Sciences'', ``Engineering'', ``Humanities \& Society'', ``Modern Architecture'', ``Ancient Architecture'', ``Geography \& Meteorological'' and ``Life Culture \& Art''). These topics encompass 56 fine-grained subtopics. 
\item \textbf{High-quality:} We implement a rigorous pipeline for the benchmark construction, including automatic verification, difficulty filtering, and human verification. 
\item \textbf{Static:} To maintain the enduring quality of ChineseSimpleVQA, all reference answers will remain unchanged over time.
\item \textbf{Easy to evaluate:} All of the questions and answers are in a short format for quick evaluation. Furthermore, we have open-sourced all datasets and code, providing ready-to-run scripts to assist researchers in their endeavors.
\end{itemize} 

Subsequently, we conducted a comprehensive evaluation of 35 advanced LVLMs, covering 8 major topics and 56 sub-topics. Here are some insightful findings:

\begin{itemize}[leftmargin=*]
\item \textbf{ChineseSimpleVQA is challenging:} The closed-source model o1-preview achieved the best performance in terms of visual factuality, surpassing the top open-source models by approximately 20 points. This indicates that open-source LVLMs still have a long way to go in this area.
\item \textbf{Larger models lead to better results:} Within the same series of models, a larger model size generally yields better results. This trend holds for nearly all series of models. 
\item \textbf{Larger models exhibit better calibration:} We observe that o1-preview is better calibrated than o1-mini, and GPT-4o is more calibrated than GPT-4o-mini.
\item \textbf{More sampling increase performance:} Increasing the sampling attempts (Best-of-N) can improve the model's performance, but this tends to stabilize after around 30 times.
\item \textbf{See the world, but not discover knowledge:} Multi-hop questions demonstrate that even the LVLMs can correctly recognize objects, the final performance may still fall short.
\end{itemize}

\section{Related Works}

\paragraph{VQA Benchmarks:}
The problem of Visual Question Answering (VQA) was originally conceptualized as a Visual Turing Test \cite{geman2015visual}. Addressing this test necessitates that computational systems replicate human faculties, including real-world visual recognition, comprehension of language, rudimentary reasoning, and factual knowledge. Certain datasets emphasize the evaluation of perceptual and linguistic understanding capabilities \cite{antol2015vqa,malinowski2014multi,goyal2017making,liu2024visual,liu2025mmbench,yu2023mm,li2024seed}, whereas others delve into the intricacies of complex reasoning processes \cite{johnson2017clevr,yue2024mmmu}. Notwithstanding these advancements, assessing factuality remains a formidable challenge, an aspect that has received limited attention within the scope of VQA datasets.

\paragraph{Factuality Benchmarks:}
Several datasets also necessitate factual or commonsense knowledge \cite{marino2019ok, wang2015explicit, wang2017fvqa, zellers2019recognition, tan2024chinese}. For instance, OK-VQA \cite{marino2019ok} evaluates a model's capability to answer questions using open-domain knowledge. Building upon this, A-OKVQA \cite{schwenk2022okvqa} integrates diverse external knowledge and reasoning capabilities, while S3VQA \cite{jain2021select} introduces a dataset featuring questions that require object detection within images. Recently, factuality benchmarks have become increasingly important for LLMs: SimpleQA introduced by OpenAI \cite{Wei2024MeasuringSF} assesses short-form factuality, and a Chinese SimpleQA \cite{he2024chinese} is proposed by Alibaba Group. 


\paragraph{Comparison to other knowledge-based VQA benchmarks:}
In Table~\ref{table.1}, we present a comparative analysis of ChineseSimpleVQA alongside several mainstream knowledge-based VQA benchmarks. Our dataset represents the first Chinese multimodal evaluation set designed to comprehensively assess factual abilities, thereby addressing a significant gap in the multimodal domain.
Moreover, the multi-hop questioning in our dataset queries the image's content and related facts, effectively evaluating the model's object recognition and factual knowledge handling.


\section{ChineseSimpleVQA}
\subsection{Overview}
As shown in Figure \ref{fig.method}, ChineseSimpleVQA is collected through a rigorous processing pipeline including automatic verification, difficulty filtering, and human verification to ensure high quality and appropriate difficulty levels. Ultimately, we obtained 1,100 images and 2,200 Q\&A pairs covering 56 knowledge subtopics. Moreover, drawing from human cognitive processes, each image is designed to include two questions: the first question is solely for visual object recognition (i.e., seeing the world), while the second question requires both object recognition and factual knowledge (i.e., discovering knowledge). In other words, these two questions are used to evaluate the model's ability to ``recognize everything" and ``understand encyclopedias'', respectively. Data examples can be found in Appendix~\ref{section:examples}.
\subsection{Knowledge Collection}
To create visually oriented Q\&A pairs based on encyclopedic knowledge, we collected a substantial number of images along with their corresponding knowledge texts from Wikipedia's ``Picture of the Day'' column\footnote{\url{https://en.wikipedia.org/wiki/Wikipedia:Picture_of_the_day}}. Notably, since our goal is to build a Chinese benchmark, we specifically used the Chinese version of the column as our data source. In total, we gathered 6,546 image-text pairs, with the text having an average length of 723 characters.

\subsection{Q\&A Construction}
\label{standard}
Our benchmark focuses on both the visual recognition and knowledge discovery capabilities of LVLMs. Therefore, we construct Q\&A pairs through three stages:
    (1). \textit{Building Image Object Recognition Questions}: Identify the main objects within the images and formulate questions.
    (2). \textit{Generating Knowledge-Based Questions on Image Objects}: 
Develop questions that investigate deeper into the knowledge related to the identified objects based on their knowledge texts.
    (3). \textit{Merging the Previous Two Steps to Create Two-Hop Visual Knowledge Questions}: Integrate the visual recognition and knowledge-based questions to create more complex, two-hop questions that are grounded in visual content.

In the final version of our dataset, we retain the Q\&A pairs from the \textbf{first and last stages} for each image to allow users to analyze the capability boundaries and execution mechanisms of LVLMs. Notably, the Q\&A pairs at all stages are initially generated by prompting the LLM\footnote{GPT-4o(0806)\cite{openai2023gpt} is used for the generation of Q\&A pairs, automatic quality verification, and evaluation.} based on the criteria outlined in the following paragraphs. Notably, all Prompts are provided in Appendix~\ref{section:prompt}.

\textbf{Criteria of Object Recognition Q\&A}. 1) Questions should be answerable using only information in the image, without needing external knowledge. For example, the question ``Who designed the poster in the image?'' is avoided as this requires extra information. 2) Each question must have a single, clear answer. For example, ``What is in the image?'' is unsuitable as it can refer to multiple entities. 3) Each question should be clear and specific, not divisible into separate inquiries.

\begin{table}[ht]
\centering
\footnotesize 

\begin{tabular}{lr}
    \toprule
    \textbf{Question Category} & \textbf{Proportion} \\
    \midrule
    \textbf{Total Number} & 2,200 \\
    Nature & 26.2\% \\
    Life, Culture \& Art & 14.5\% \\
    Geography \& Meteorological & 12.4\% \\
    Modern Architecture & 11.8\% \\
    Human \& Society & 9.8\% \\
    Ancient Architecture & 9.7\% \\
    Sciences & 8.4\% \\
    Engineering & 7.2\% \\
    \midrule
    \textbf{Recognition Question Length} &  \\
    \midrule
    - \textit{Maximum Length} & 28 \\
    - \textit{Minimum Length} & 6\\
    - \textit{AVG Length} & 11.1 \\
    \midrule
    \textbf{Final Question Length} &  \\
    \midrule
    - \textit{Maximum Length} & 46 \\
    - \textit{Minimum Length} & 8 \\
    - \textit{AVG Length} & 16.1 \\
    \midrule
    \textbf{Recognition Answer Length} & \\
    \midrule
    - \textit{Maximum Length} & 16 \\
    - \textit{Minimum Length} & 1 \\
    - \textit{AVG Length} &  5.2\\
    \midrule
    \textbf{Final Answer Length} & \\
    \midrule
    - \textit{Maximum Length} & 33\\
    - \textit{Minimum Length} & 1 \\
    - \textit{AVG Length} &  4.5\\
    \bottomrule
  \end{tabular}
\caption{Dataset statistics.}
\vspace{-0.4cm}
\label{tab:state}
\end{table}

\textbf{Criteria of Knowledge-Based Q\&A}. 1) Questions should rely on factual knowledge, avoiding subjective opinions or personal views.  2) Answers must be clear and distinct, similar to the previous point.  3) Do not ask two questions at once, consistent with the previous guideline.  4) Answers should be timeless and not change over time, reflecting enduring facts. 5) Questions should be challenging, and in-depth knowledge should be assessed, avoiding overly simple queries. 6) Keep questions concise by including only essential information for a clear answer.  For example, instead of asking, ``In which country is the Desen'ka train station located in Vinnytsia City?'' ask, ``In which country is the Desen'ka train station located?''







\textbf{Criteria of Final Merged Q\&A}. 1) Questions must rely entirely on the answer to the first Q\&A pair, meaning the next question cannot be answered correctly without the first answer.2) Ensure that new questions are clear and concise, maintaining fluency and grammatical accuracy.


\subsection{Quality Control}
We design strict quality control processes, including automatic verification and human annotation, to identify and correct corresponding issues. 

\paragraph{Automatic Verification.} We developed a multi-round automated method to verify the quality of synthetic Q\&A pairs, aiming to regenerate or filter out questions that do not meet the above criteria.  Upon completion of the automatic verification, we retained a total of 5,009 images with 10,018 qualified Q\&A pairs.

\paragraph{Difficulty Filtering.} Subsequently, we filtered out simple samples to identify the knowledge boundaries of LLMs and enhance the difficulty of the benchmark. Specifically, if a question can be correctly answered by all of the four powerful models, i.e., GPT-4o (0806), Claude 3.5 Sonnet \cite{enis2024llm}, Gemini 1.5 Pro \cite{team2024gemini}, and Qwen-VL-Max \cite{bai2023qwen}, it is considered simple and thus discarded. As a result of this process, a total of 3,058 images with 6,116 Q\&A pairs were retained.

\paragraph{Human Verification.} 
During this parse, a team of 23 annotators performed data verification and rewriting, while 6 engineers reviewed and selected the high-quality data. 

Annotators must ensure question-answer pairs meet the standards by the following operations: 1) verifying and rewriting the questions or answers that do not meet the standards in Section\ref{standard}; 2) replacing unqualified or unrepresentative images with new ones from online sources; and 3) verifying answer accuracy via search engines and Baidu Baike. Pairs that cannot be improved are discarded. 
Finally, manual annotations are compared with LLM-verified results, retaining only fully consistent Q\&A pairs. This rigorous process ensures accuracy and adherence to established standards.

\paragraph{Data Desensitization.}
To mitigate potential security risks during evaluation, we submitted the final dataset to six security auditors for a comprehensive security review. Each sample of data was cross-checked by at least two auditors, and only the data that passed all scrutiny was retained.

After the above annotations, 2,411 images with 4,822 Q\&A pairs were retained. Afterward, the algorithm experts carefully selected 1,100 images with 2,200 QA pairs as the final dataset.

\subsection{Dataset Statistics}
As shown in Table~\ref{tab:state}, ChineseSimpleVQA consists of 8 major topics and 56 subtopics, which are based on manually labeled tags. Among these topics, natural knowledge is the most common, comprising 26.2\% of the total. Furthermore, since ChineseSimpleVQA primarily evaluates short-form answers, the average answer lengths for these two stages of questions are 5.2 words and 4.5 words, respectively, enabling a quicker evaluation.


\begin{table*}[ht]
\centering
\small
\begin{tabular}{lccccc|ccccc}
\toprule
\multirow{2}{*}{\textbf{Model}} & \multicolumn{5}{c}{\textbf{Overall results of Merged Q\&A}} & \multicolumn{5}{c}{\textbf{Overall results of Recognition Q\&A}} \\
\cmidrule(r){2-6} \cmidrule(lr){7-11}
& CO & IN$\downarrow$ & NA$\downarrow$ & CGA & F-score & CO & IN$\downarrow$ & NA$\downarrow$ & CGA & F-score \\
\midrule
\multicolumn{11}{c}{\textit{Closed-Source Large Vision Language Models}} \\
\midrule
o1-preview & \colorbox{firstBest}{\textbf{68.8}} & \colorbox{firstBest}{\textbf{24.6}} & 6.5 & \colorbox{firstBest}{\textbf{73.6}} & \colorbox{firstBest}{\textbf{71.1}} & \colorbox{firstBest}{\textbf{79.1}} & \colorbox{firstBest}{\textbf{13.6}} & 7.3 & \colorbox{firstBest}{\textbf{85.3}} & \colorbox{firstBest}{\textbf{82.1}} \\
o1-mini & 52.7 & 38.1 & 9.2 & 58.1 & 55.3 & 64.8 & 24.9 & 10.3 & 72.2 & 68.3  \\
GPT-4o & 59.1 & 35.5 & 5.4 & 62.4 & 60.7 & 77.5 & 15.5 & 7.0 & 83.4 & 80.4  \\
GPT-4o-mini & 51.0 & 43.6 & 5.4 & 53.9 & 52.4 & 70.8 & 23.1 & 6.1 & 75.4 & 73.0\\
Gemini-2.0-flash & 64.5 & 29.5 & 5.9 & 68.6 & 66.5 & 76.7 & 19.6 & 3.7 & 79.7 & 78.2  \\
Gemini-1.5-pro-flash & 56.5 & 34.6 & 8.8 & 62.0 & 59.2 & 70.3 & 25.9 & 3.8 & 73.1 & 71.6 \\
Gemini-1.5-pro & 66.2 & 31.4 & \colorbox{firstBest}{\textbf{2.5}} & 67.8 & 67.0 & 77.5 & 20.7 & \colorbox{firstBest}{\textbf{1.7}} & 78.9 & 78.2 \\
Claude-3.5-sonnet2 & 63.8 & 30.6 & 5.5 & 67.6 & 65.6 & 77.6 & 17.2 & 5.2 & 81.9 & 79.7 \\
Claude-3.5-sonnet & 59.5 & 26.4 & 14.2 & 69.4 & 64.0 & 69.5 & 20.2 & 10.3 & 77.5 & 73.3 \\
Qwen-VL-max & 56.5 & 39.6 & 3.8 & 58.8 & 57.6 & 72.9 & 24.6 & 2.5 & 74.7 & 73.8  \\
Doubao-1.5-vision-pro & 53.2&	36.4&	10.5&	59.4&	56.1&	50.4&	39.1&	10.5&	56.3&	53.2\\
Doubao-vision-pro & 52.0&	43.2&	4.8&	54.6&	53.3&	44.0&	51.9&	4.1&	45.9&	44.9 \\
Doubao-vision-lite & 31.9&	30.0&	38.1&	51.5&	39.4&	30.6&	43.5&	25.8&	41.3&	35.2\\
\midrule
\multicolumn{11}{c}{\textit{Open-Source Large Vision Language Models}} \\
\midrule
Deepseek-VL2 & 33.4 & 60.3 & 6.4 & 35.6 & 34.5 & 32.0&	58.5&	9.5&	35.4&33.6\\
Deepseek-VL2-small & 37.0 & 58.5 & 4.5 & 38.8 & 37.9 & 33.1 &	58.4 &	8.5 &	36.2 &	34.6 \\
Deepseek-VL2-tiny & 22.5 & 72.3 & 5.2 & 23.8 & 23.1 & 23.0& 	69.9& 	7.1& 	24.8& 	23.8\\
\hline
LLaVA-onevison-72B & 34.7 & 60.6 & 4.6 & 36.4 & 35.6 & 29.5&	59.6&	10.9&33.1&	31.2 \\
LLaVA-onevison-7B & 18.0&	57.7&	24.3&	23.8&	20.5&	14.5&	61.2&	24.4&	19.1&	16.5\\
LLaVA-onevison-0.5B & 7.9 & 82.3 & 9.8 & 8.8 & 8.3 & 8.5 & 70.6 &	20.8 &10.8 &	9.5  \\
\hline
Qwen2.5-VL-72B&	49.0&	\colorbox{firstBest}{\textbf{42.8}}&	8.2&	\colorbox{firstBest}{\textbf{53.4}}&	51.1&	45.7&	\colorbox{firstBest}{\textbf{48.5}}&	5.7&	48.5&	47.1\\
Qwen2.5-VL-7B&	39.5&	54.2&	6.4&	42.1&	40.8&	41.3&	51.2&	7.5&	44.6&	42.9\\
Qwen2.5-VL-3B&	31.4&	61.3&	7.4&	33.9&	32.6&	32.9&	62.0&	5.1&	34.7&	33.8\\

Qwen2-VL-72B&	\colorbox{firstBest}{\textbf{50.6}}&	46.6&	2.7&	52.1&	\colorbox{firstBest}{\textbf{51.3}}&	\colorbox{firstBest}{\textbf{48.0}}&	49.1&	\colorbox{firstBest}{\textbf{2.9}}&	\colorbox{firstBest}{\textbf{49.4}}&	\colorbox{firstBest}{\textbf{48.7}}\\
Qwen2-VL-7B&	38.2&	60.3&	\colorbox{firstBest}{\textbf{1.5}}&	38.8&	38.5&	39.5&	57.1&	3.4&	40.9&	40.2\\
Qwen2-VL-2B	&29.0&	66.5&	4.5&	30.4&	29.7&	34.2&	63.1&	2.7&	35.1&	34.7\\
\hline

GLM-4v	&34.8&	51.7&	13.5 &	40.2&	37.3&	37.1&	54.5&	8.4&	40.5&	38.7 \\
\hline

Llama-vision-90B&	46.2&	51.6&	2.2&	47.2&	46.7&	37.2&	57.5&	5.3&	39.3&	38.2\\
Llama-vision-11B&	29.4&	66.4&	4.3&	30.7&	30.0&	27.4&	64.3&	8.4&	29.9&	28.6\\
\hline
InterVL2.5-78B	&41.1&	51.7&	7.2&	44.3&	42.6&	36.2&	56.7&	7.1&	38.9&	37.5\\
InterVL2.5-38B	&35.5&	54.3&	10.2&	39.6&	37.5&	30.0&	56.4&	13.6&	34.7&	32.2\\
InterVL2.5-26B	&32.4&	52.9&	14.7&	38.0&	34.9&	29.9&	55.3&	14.8&	35.1&	32.3\\
InterVL2.5-8B	&28.4&	60.5&	11.1&	31.9&	30.0&	24.5&	63.5&	12.1&	27.8&	26.0\\
InterVL2.5-4B	&24.5&	73.7&	1.8&	24.9&	24.7&	22.2&	72.5&	5.4&	23.4&	22.8\\
InterVL2.5-2B	&13.9&	73.9&	12.2&	15.8&	14.8&	16.8&	74.0&	9.2&	18.5&	17.6\\
InterVL2.5-1B	&14.7&	70.7&	14.5&	17.2&	15.9&	16.8&	69.3&	13.9&	19.5&	18.1\\
\bottomrule
\end{tabular}
\vspace{-0.2cm}
\caption{Performance comparison of closed source and open source LVLMs on multi-hop QAs (i.e. Merged Q\&A and Recognition Q\&A). For metrics, \textbf{CO}, \textbf{NA}, \textbf{IN}, and \textbf{CGA} denote ``Correct'', ``Not attempted'', ``Incorrect'', and ``Correct given attempted'', respectively. The highest scores among models in each metric are highlighted in \colorbox{firstBest}{\textbf{green}}.}
\label{tab:model-performance}
\vspace{-0.4cm}
\end{table*}

\section{Experiment}

\subsection{Setup}
We maintain a consistent prompt format throughout all experiments, ensuring uniformity in our benchmark. The temperature and sampling parameters are set to the official default settings for each model. For evaluation, we use the ``GPT-4o-0806'' as a judge model.

\subsection{Baseline Models}
We evaluate \textbf{13 closed-source LVLMs} (i.e., o1-preview (0901)
, o1-mini (0901)
, GPT-4o (0806)
, GPT-4o-mini (0708)
\footnote{\url{https://openai.com/index/}}
, Gemini-1.5-pro-flash, Gemini-1.5-pro \cite{team2024gemini}, Gemini-2.0-flash\footnote{\url{https://deepmind.google/technologies/gemini/flash/}}, Claude-3.5-sonnet2\footnote{\url{https://www.anthropic.com/news/claude-3-5-sonnet}}, Claude-3.5-sonnet, Doubao-1.5-vision-pro\footnote{\url{https://team.doubao.com/en/direction/vision}}, Doubao-vision-pro/lite, Qwen-VL-max \cite{bai2023qwen}) and \textbf{22 open-source LVLMs} (i.e., Qwen2-VL series \cite{wang2024qwen2}, Qwen2.5-VL series\footnote{\url{https://help.aliyun.com/zh/model-studio/developer-reference/use-qwen-by-calling-api}},Deepseek-VL-2 series \cite{wu2024deepseek}, LLaVA-one-vision-0.6B/72B \cite{li2024llava}, GLM-4V-9B\footnote{\url{https://open.bigmodel.cn/dev/api/normal-model/glm-4v}}, Llama-vision-11B/90B\footnote{\url{https://ollama.com/library/llama3.2-vision}}, InternVL series \cite{chen2024internvl}.).

\subsection{Evaluation Metrics}
\label{metric}
We use the following metrics to evaluate ChineseSimpleVQA:
(1) \textbf{Correctness (CO)}: The predicted answer to the final question should fully encompass the reference answer without contradictions.
(2) \textbf{Not attempted answers(NA)}: The predicted answer to the final question is not related to the reference answer without any contradictions.
(3) \textbf{Incorrectness (IN)}: The predicted answer to the final question contradicts the reference answer, as does the answer to the first question.
(4) \textbf{Correctness among Given Attempted (CGA)}: The proportion of questions accurately answered among those attempted.
(5) \textbf{F-score}: The harmonic mean of Correct and Correct Given Attempted.

\begin{figure*}[t]
\begin{center}
\includegraphics[scale=0.21]{combined_dual_radar_charts1.png}
\caption{Correctness(CO) metric for eight topics. To conserve space, only the top 10 models are displayed.}
\label{fig.topics}
\end{center}
\vspace{-0.6cm}
\end{figure*}
\begin{figure}[t]
\begin{center}
\includegraphics[scale=0.4]{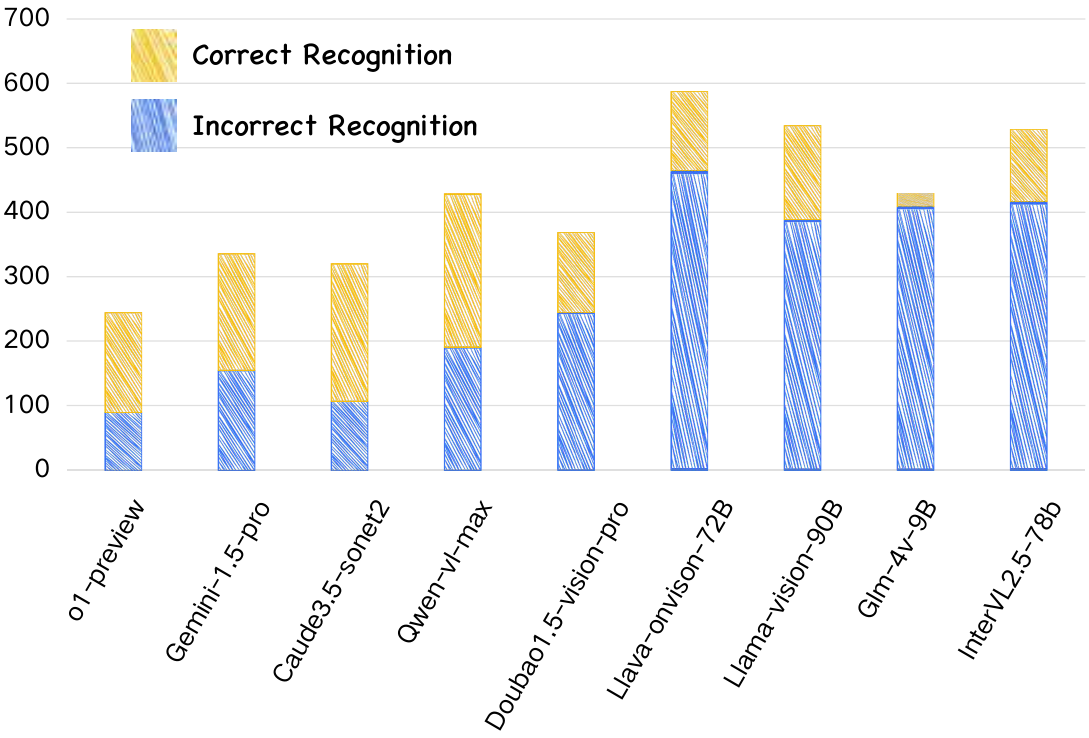}
\caption{The number of incorrectly answered final questions, and the number of samples where the image object was correctly recognized.}

\label{fig.error}
\end{center}
\vspace{-0.6cm}
\end{figure}
\begin{figure*}[t]
\begin{center}
\includegraphics[scale=0.3]{calibration_and_inference_comparison.png}
\caption{Up: Calibration of LLMs based on their stated confidence for Recognition and final Q\&A. Down: Improvement in
accuracy with increased test-time compute using Best-of-N for Recognition and final Q\&A.}
\label{fig.calibration}
\end{center}
\vspace{-0.6cm}
\end{figure*}

\subsection{Main Results}\label{sec:main}
We present the performance results of various LVLMs on the ChineseSimpleVQA dataset in Table~\ref{tab:model-performance}. The evaluation metrics are introduced in Section~\ref{metric}. We report the models' performance on recognition and final merged questions, and summarize the following findings:
\begin{itemize}[leftmargin=*]
    \item \textbf{o1-preview is the best} performer for both recognition and final questions among all models. The next best are Gemini-2.0-flash and Gemini-1.5-pro, which follow closely behind. For Chinese ability, Qwen family performs the best.
    \item \textbf{Performance on final questions is positively correlated with the performance on recognition questions}, though accuracy is typically higher for recognition questions. This suggests that models can identify objects but sometimes fail to grasp deeper knowledge. Some models (e.g. Deepseek-VL2) perform better on final questions because correct and incorrect object predictions can point to the same final answer. For instance, both ``Misumena vatia'' and ``Thomisidae'' belong to the order ``Araneae.''\footnote{Examples are the scientific names of two spider species.}.
    \item \textbf{The larger the model size within the same series of models, the better the result.} Taking the Qwen2-VL series as an example, when the model size increases from 3B to 72B, the accuracy on the final questions increases from 29.0\% to 50.6\%. More detailed analysis can be found in Appendix \ref{app:scale}.
    \item \textbf{The IN rate of the models is much higher than their NA rate} (in addition to Doubao-vision-lite), indicating that the models are more likely to confidently provide incorrect factual information. Mitigating this hallucination problem remains a significant challenge. In Section \ref{ana}, we will conduct a detailed analysis focusing on model confidence.
\end{itemize}

\subsection{Performance on Different Topics}
We select the top 10 models with the highest CO rate on the final questions and evaluated their performance across different topics. The results are shown in Figure \ref{fig.topics}. For all topics, the trends in model performance for both recognition and final questions align with the conclusions presented in the previous section. It is obvious that these models are more adept at topics such as modern architecture, engineering, and science. However, their performance is slightly lower on natural topics, as these questions often need more specialized knowledge than common sense can provide.

From this figure, we can observe the differences in the model's performance on recognition questions versus final questions across different topics. For example, Claude-3.5-Sonnet exhibits a small performance gap in the topic of modern architecture. This indicates that once the model accurately identifies the object of an image, it is more likely to provide accurate responses based on relevant knowledge. Conversely, if the model demonstrates a large performance gap on a particular topic, it suggests that the model's knowledge of that topic is limited.

\subsection{Further Analysis}
\label{ana}
\paragraph{Analysis of Knowledge Scope.}
In the design of ChineseSimpleQA, the final question requires deeper, extended knowledge beyond the recognition question. To explore at which stage the real knowledge limitation arises when models fail to answer the final question, we collect examples where different models provided incorrect answers to the final questions and conduct analysis by calculating the percentage of corresponding recognition questions that were answered correctly. The results are presented in Figure \ref{fig.error}.

For o1-preview, Gemini-1.5-pro, Caude3.5-sonet2 and Qwen-vl-max, the error rate caused by image object recognition is less than 50\%. This shows that for these models, the inability to correctly answer the final question is mainly limited by the lack of more complex extended knowledge. For the other models, it is impossible to correctly recognize the image object from the initial stage, especially GLM-4v-9B.

\paragraph{Analysis of Calibration.}
The confidence level (\%) of a perfectly calibrated model should precisely match the accuracy of its predictions. From this perspective, we prompt the models to provide their confidence levels (ranging from 0 to 100) while answering questions. In Figure \ref{fig.calibration}, we present the accuracy of the models' answers at different confidence intervals (with an interval size of 10).

The results illustrate the alignment performance of the tested models. Specifically, o1-preview exhibits the best performance on both types of questions. While o1-mini ranks second in recognition questions, closely following o1-preview, it performs slightly worse on final questions. Overall, most models fall below the perfect alignment line, indicating that they tend to be overconfident, even when their answers are incorrect.
\begin{figure}[t]
\begin{center}
\includegraphics[scale=0.24]{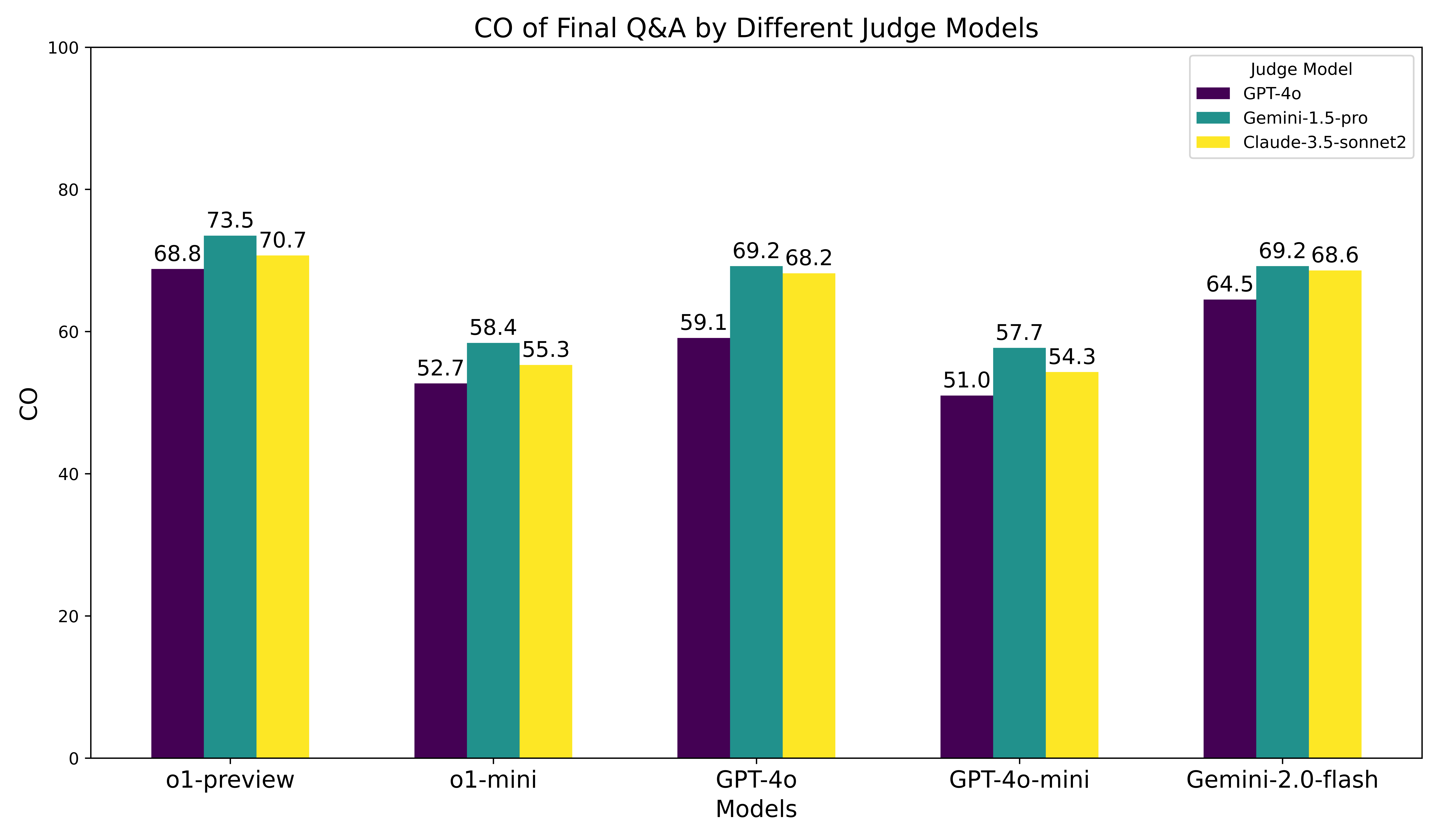}
\caption{Robustness of Judge Models of Final Q\&A.}
\label{fig.judge}
\end{center}
\vspace{-0.6cm}
\end{figure}

\paragraph{Analysis of Inference Attempt.}
Furthermore, we analyze the changes in the accuracy of the model's answers as the inference attempts increased. Specifically, we sample 100 instances from each of the two types of questions in ChineseSimpleVQA. For each instance, the model is asked to independently generate an answer 100 times. For the results of multiple inferences, we use the Best-of-N method to obtain the accuracy of the model responses.

Figure \ref{fig.calibration} shows that within the range of 1-30 inference times, the accuracy of the tested model increases with the increase of inference times. However, when the number of inferences is greater than 30, the performance of the model tends to be stable. Beyond this point, the model does not acquire significantly more accurate knowledge even with additional attempts. These findings can serve as a reference for determining the boundaries of the model’s knowledge capabilities.

\section{Robustness of Judge Models}
Moreover, we select two models,i.e.,Gemini-1.5-pro and Claude-3.5-sonnet2, as the judge models and compare the evaluation results of GPT-4o-0806. The results are shown in Figure \ref{fig.judge}. It can be seen that when using different evaluation models, the ranking of the tested models remains consistent. This further proves the robustness of the evaluation method of ChineseSimpleVQA.

\section{Conclusion}
We introduce the first short-form VQA benchmark, named Chinese SimpleVQA, to evaluate the visual factuality abilities of existing LVLMs. This benchmark is characterized by important features such as multi-hop, diverse, high-quality, static, and easy-to-evaluate. Based on Chinese SimpleVQA, we conduct a comprehensive evaluation of the performance of nearly 30 models regarding visual factuality, providing a detailed analysis of the capability boundaries to demonstrate the advantages and necessity of ChineseSimpleVQA.

\section{Limitation}
From the perspective of topic distribution, the coverage of topics discussed in this paper remains relatively limited. A broader topic coverage would contribute to a more comprehensive assessment of visual knowledge. In terms of data volume, after rigorous quality filtering and difficulty balancing, the final version of our dataset represents only a tiny fraction of the overall data. Therefore, expanding the dataset will be a key focus of our future efforts. Furthermore, as highlighted by this work \cite{jiang2025hiddendetect}, datasets to evaluate the model against potential jailbreak attacks is crucial. Finally, regarding model structure, improving the visual factuality of the model to enhance its performance on ChineseSimpleVQA will become increasingly important. However, both performance improvement and security aspects may already be beyond the scope of this paper.

\section{Ethic Statement}
We respect intellectual property rights and comply with relevant laws and regulations. The source of our dataset is publicly available, and we have taken careful measures to ensure that the dataset does not contain any personal sensitive information. In addition, our work is only for research purposes, not for commercial purposes.

\bibliography{custom}

\appendix

\section{Overview of Appendix}
We have over 10 pages of this appendix, comprising the following subsections for the convenience of readers:

\begin{figure*}[t]
\begin{center}
\includegraphics[scale=0.6]{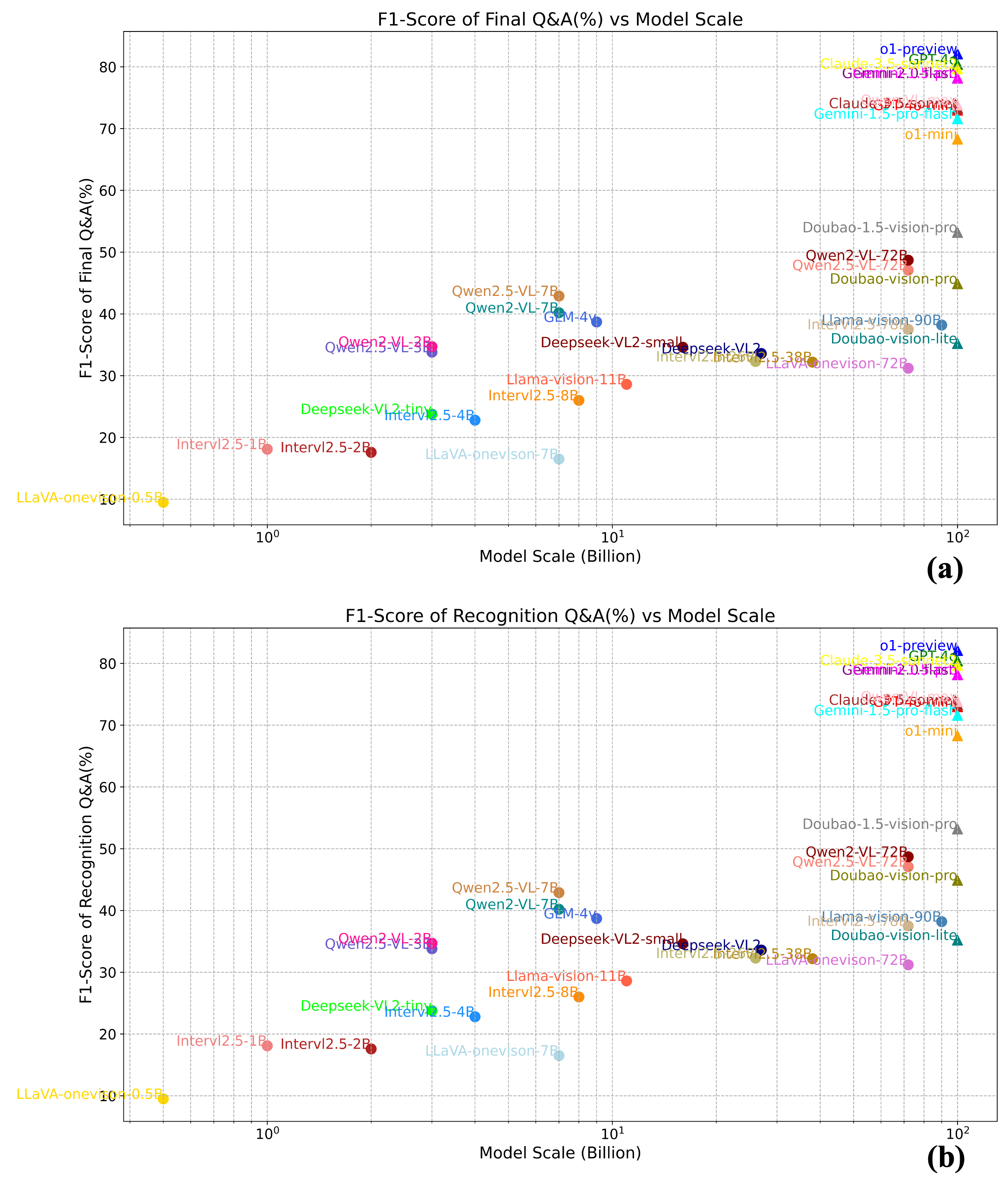}
\caption{Relationship between model scale (in billion
parameters) and F-score on Recognition and final Q\&A.}
\label{fig.scale}
\end{center}
\vspace{-0.6cm}
\end{figure*}
\begin{figure}[ht]
\begin{center}
\includegraphics[scale=0.4]{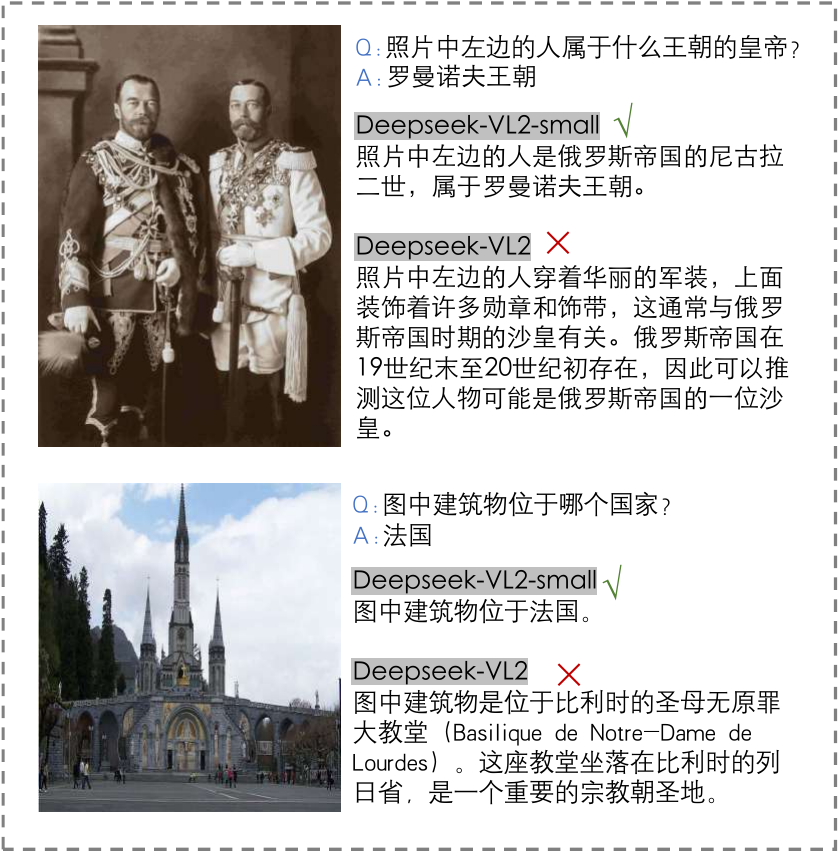}
\caption{Example of model responses of where Deekseep-vl2-small's answer is correct but Deepseek-VL2's answer is wrong.}
\label{fig.deepseek}
\end{center}
\vspace{-0.6cm}
\end{figure}
\begin{figure*}[t]
\begin{center}
\includegraphics[scale=0.21]{combined_dual_radar_charts_CGA1.png}
\caption{CGA of the top 10 models for eight topics.}
\label{fig.CGA}
\end{center}
\end{figure*}

\begin{figure*}[t]
\begin{center}
\includegraphics[scale=0.24]{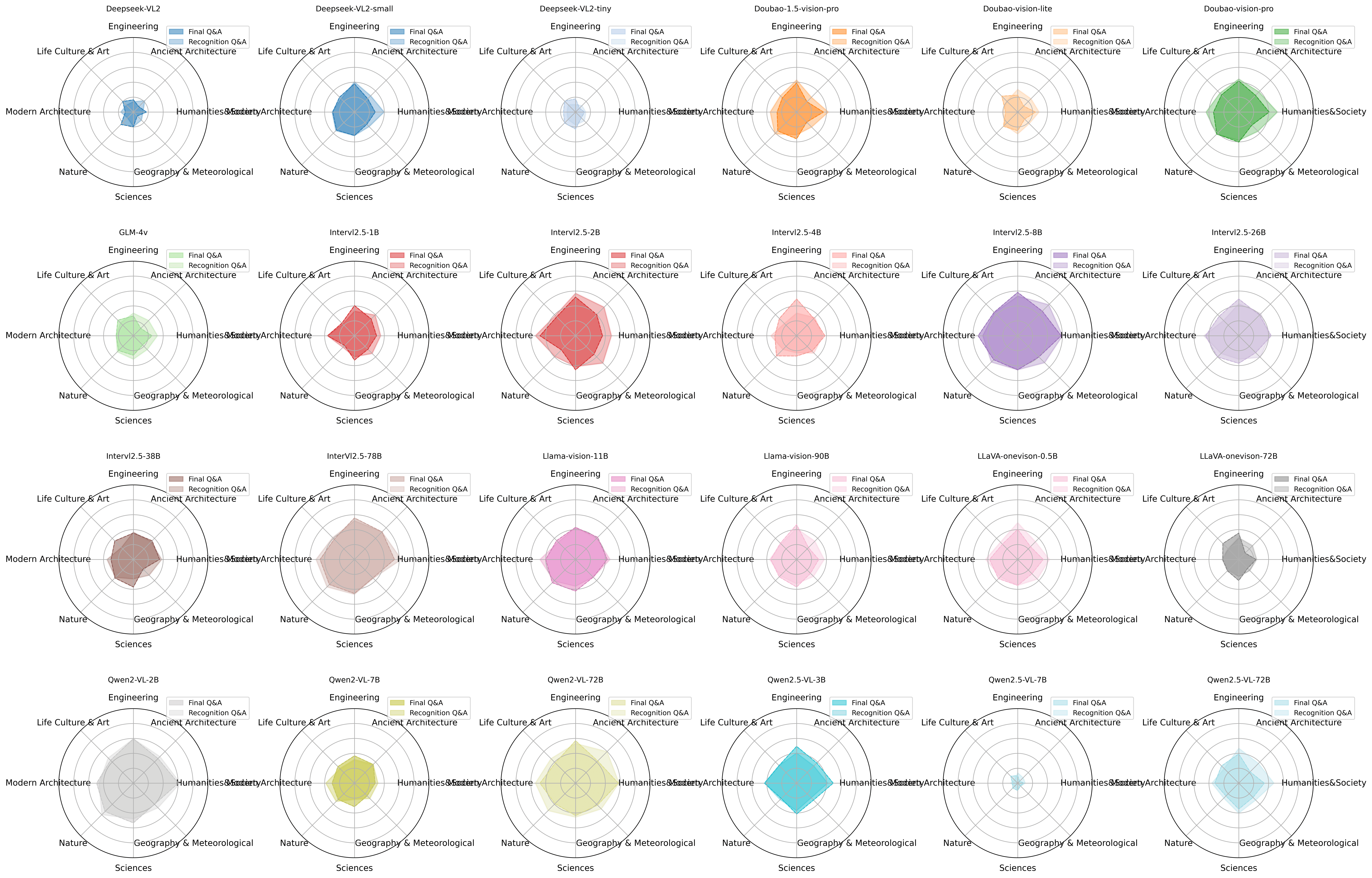}
\caption{CO of all other different models for eight topics.}
\label{fig.co1}
\end{center}
\vspace{-0.6cm}
\end{figure*}

\begin{figure*}[t]
\begin{center}
\includegraphics[scale=0.24]{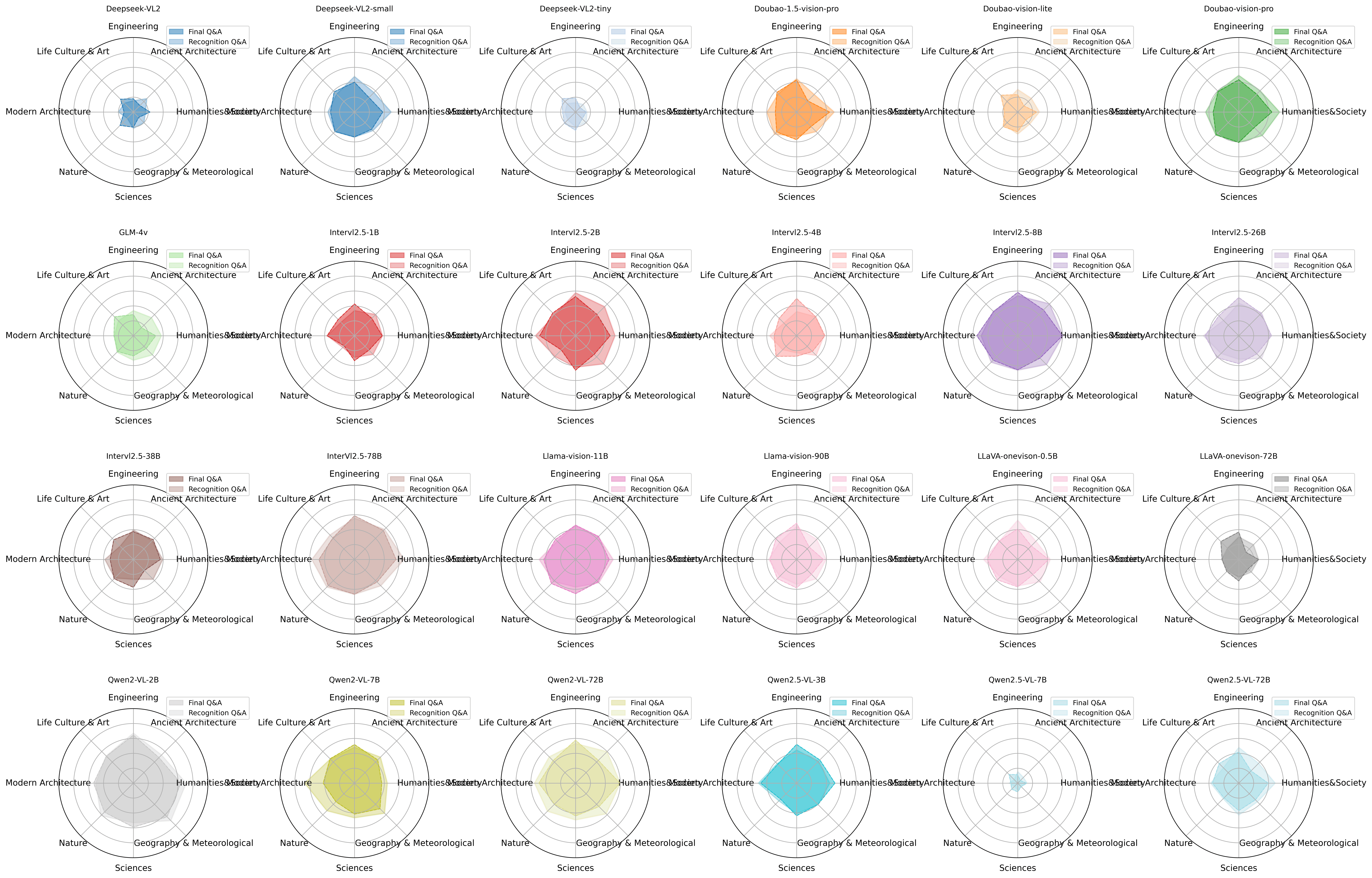}
\caption{CGA of all other different models for eight topics.}
\label{fig.CGA1}
\end{center}
\vspace{-0.6cm}
\end{figure*}

\noindent \textbf{More additional experiments}
\begin{itemize}[leftmargin=*]
    \item \textbf{Appendix \ref{app:scale}}: Additional experimental analysis of model scale.
    \item \textbf{Appendix \ref{app:cga}}: Additional experimental analysis of performance comparison among models.
    \item \textbf{Appendix \ref{app:rank}}: Additional experimental analysis of rankings on ChineseSimpleVQA.
    \item \textbf{Appendix \ref{app:Popularity}}: Additional experimental analysis between popularity and difficulty.
    \item \textbf{Appendix \ref{app:Human}}: Additional analysis of Human Calibration.
    \item \textbf{Appendix \ref{app:other}}: Additional experimental comparion with other Datasets.
    \item \textbf{Appendix \ref{app:Text}}: Additional experimental comparison with Knowledge-Based Q\&A
\item \textbf{Appendix \ref{section:more}}:
 More results.
\end{itemize}

\noindent \textbf{More visualization and cases}
\begin{itemize}[leftmargin=*]
  \item \textbf{Appendix \ref{section:len}}: Detailed question distribution.
    \item \textbf{Appendix \ref{section:examples}}: Dataset visualization of our ChineseSimpleVQA.
    \item \textbf{Appendix \ref{app:response}}: Visualization of performance comparison among models.
    \item \textbf{Appendix \ref{section:prompt}}: Details of all prompts in the dataset generation and validation.
\end{itemize}

We hope that our efforts will serve as a source of inspiration for more readers!

\section{More Experiments}

\subsection{F-score v.s. Model Scale}\label{app:scale}
In Figure \ref{fig.scale}, we illustrate the trend of F-score as the model size increases. Generally, there is a positive correlation between these two factors. However, this relationship weakens in the intermediate range, especially for vision recognition tasks, where larger models may perform worse.
This observation contrasts with findings from ChineseSimpleQA, which analyzed large language models (LLMs). For large vision-language models (LVLMs), while increased size suggests greater knowledge capacity, accurate image object recognition hinges more on the quality of image-text alignment training, critically affecting factual accuracy.
Additionally, Table \ref{tab:model-performance} shows the performance of same-series models typically improves with size, but Deepseek-VL and Deepseek-VL-small deviate from this pattern. Figure \ref{fig.deepseek} provides examples where the smaller model correctly answers the questions that the larger Deepseek-VL fails, highlighting the discrepancy between them.

\subsection{CGA Results on 8 Topics}\label{app:cga}
In this section, we present more results of Correctness among Given Attempted (CGA). In principle, the models are encouraged to do not attempt to answer when they are unable to determine the knowledge. Therefore, we further show the accuracy of the models' responses on 8 topics after excluding the questions that the models do not attempt to answer, namely CGA. In general, GCA has higher indicators than CO. The detail results are shown in Figure \ref{fig.CGA}.

\subsection{Rankings on ChineseSimpleVQA}\label{app:rank}
We also compare the ranking differences of various models on the recognition questions and the final questions. The rankings are shown in Figure \ref{fig.rank}.

\subsection{Analysis between Popularity and Difficulty}\label{app:Popularity}
We select the first 100 samples from our dataset and monitor their page views on Wikipedia over the past 30 days to assess their popularity. We categorize the popularity into three ranges based on the number of views: (0-10), (10-50), and (50, +inf). We then evaluate the experimental performance of the following 10 closed-source models across these different levels of popularity as shown in Table \ref{tab:popularity}. The results show that models generally perform better with higher popularity data, particularly in Recognition Q\&A tasks. For data with over 50 views in the past 30 days, most models can achieve an accuracy rate of 100\%.

\subsection{Analysis of Human Calibration}\label{app:Human}
We manually extract 110 samples of the main results from both the merged and recognition Q\&A for verification and find that agreement with human assessments reached 95.5\% and 97.3\%, respectively. This proves the validity of using LLM-as-judge for evaluation.

\subsection{Comparion with other Datasets}\label{app:other}
In order to explore whether models that perform well in other datasets continue to perform well in this dataset. We compare the evaluation performance of models on our ChineseSimpleVQA and text-based (no vision) short-form factuality benchmarks——ChineseSimpleQA \cite{he2024chinese}. The experimental results are shown in Table \ref{tab:chinesesipleqa}:
As can be seen from the CO metric, the o1-preview model achieves the best results in the merged Q\&A and Recognition Q\&A of ChineseSimpleVQA, as well as in ChineseSimpleQA. In contrast, the gpt-4o-mini model performs the worst in the merged Q\&A of ChineseSimpleVQA and in ChineseSimpleQA, demonstrating consistency. For the Recognition Q\&A of ChineseSimpleVQA, the model that performed the worst was o1-mini. This indicates that our benchmark is a qualified factuality evaluation set that can also assess large vision language models' image recognition capabilities. It holds broad evaluative applications for LVLMs.

\subsection{Performance Comparison including Knowledge-Based Q\&A }\label{app:Text}
To assist us in analysis, we conduct an evaluation on text-only Knowledge-Based Q\&A. The experimental results are shown in Table \ref{tab:text-only}.
From the CO metric perspective, we can find that the o1 preview still shows the best performance, which is consistent with the other two Q\&A tasks. Additionally, the performance of Knowledge-Based Q\&A is similar to Merged Q\&A, and for most models, Knowledge-Based Q\&A performs better. This indicates that for LVLMs, it is more challenging to simultaneously recognize images as well as answer Knowledge-Based questions, and compared to image recognition, Knowledge-Based Q\&A is the more difficult part.

\begin{table*}[ht]
\centering
\small
\begin{tabular}{lccc|ccc}
\toprule
\multirow{2}{*}{\textbf{Model}} & \multicolumn{3}{c}{\textbf{CO of Merged Q\&A}} & \multicolumn{3}{c}{\textbf{CO of Recognition Q\&A}} \\
\cmidrule(r){2-4} \cmidrule(lr){5-7}
& 0 - 10 & 10 - 50 & 50 - +inf &  0 - 10 & 10 - 50 & 50 - +inf \\
\midrule
o1-preview & 68.00&	84.62&	80.00& 84.00&	90.38&	100.00\\
o1-mini & 68.00&	73.08&	60.00&72.00&	76.92&	80.00 \\
GPT-4o & 72.00&	73.08&	80.00&  76.00	&86.54&	100.00\\
GPT-4o-mini &48.00&	69.23&	80.00&68.00&	92.31&	80.00\\
Gemini-2.0-flash & 72.00&	78.85&	60.00&84.00&	90.38&	80.00  \\
Gemini-1.5-pro-flash & 80.00&	71.15&	70.00 &88.00&	78.85&	80.00\\
Gemini-1.5-pro & 80.00&	76.92&	100.00& 92.00&	86.54&	100.00\\
Claude-3.5-sonnet2 & 84.00&	73.08&	80.00 &88.00&	84.62&100.00\\
Claude-3.5-sonnet & 68.00&	67.31&	80.00& 92.00&	78.85&	90.00\\
Qwen-VL-max & 88.00&	73.08&	90.00& 80.00&	78.85&	90.00 \\
\midrule
\end{tabular}
\vspace{-0.2cm}
\caption{Different popularity's difficulty of closed source LVLMs on multi-hop QAs (i.e. Merged Q\&A and Recognition Q\&A). }
\label{tab:popularity}
\vspace{-0.4cm}
\end{table*}

\begin{table*}[ht]
\centering
\tiny
\setlength{\tabcolsep}{5pt}
\begin{tabular}{lccccc|ccccc|ccccc}
\toprule
\multirow{2}{*}{\textbf{Model}} & \multicolumn{5}{c}{\textbf{Overall results of Merged Q\&A}} & \multicolumn{5}{c}{\textbf{Overall results of Recognition Q\&A}}& \multicolumn{5}{c}{\textbf{Overall results of ChineseSimpleQA}} \\
\cmidrule(r){2-6} \cmidrule(lr){7-11}\cmidrule(lr){12-16}
& CO & IN$\downarrow$ & NA$\downarrow$ & CGA & F-score & CO & IN$\downarrow$ & NA$\downarrow$ & CGA & F-score & CO & IN$\downarrow$ & NA$\downarrow$ & CGA & F-score\\
\midrule
o1-preview & \colorbox{firstBest}{\textbf{68.8}} & \colorbox{firstBest}{\textbf{24.6}} & 6.5 & \colorbox{firstBest}{\textbf{73.6}} & \colorbox{firstBest}{\textbf{71.1}} & \colorbox{firstBest}{\textbf{79.1}} & \colorbox{firstBest}{\textbf{13.6}} & 7.3 & \colorbox{firstBest}{\textbf{85.3}} & \colorbox{firstBest}{\textbf{82.1}}&\colorbox{firstBest}{\textbf{63.8}}&	\colorbox{firstBest}{\textbf{24.0}}&	12.2&	\colorbox{firstBest}{\textbf{72.7}}&	\colorbox{firstBest}{\textbf{67.9}} \\
o1-mini & 52.7 & 38.1 & 9.2 & 58.1 & 55.3 & 64.8 & 24.9 & 10.3 & 72.2 & 68.3  &39.5&	39.9&	20.6&	49.7&	44.1\\
GPT-4o & 59.1 & 35.5 & 5.4 & 62.4 & 60.7 & 77.5 & 15.5 & 7.0 & 83.4 & 80.4&59.3&	39.3&	1.4&	60.1&	59.7  \\
GPT-4o-mini & 51.0 & 43.6 & 5.4 & 53.9 & 52.4 & 70.8 & 23.1 & 6.1 & 75.4 & 73.0&37.6&	61.5&	\colorbox{firstBest}{\textbf{0.9}}&	37.9&	37.8\\
Gemini-1.5-pro & 66.2 & 31.4 & \colorbox{firstBest}{\textbf{2.5}} & 67.8 & 67.0 & 77.5 & 20.7 & \colorbox{firstBest}{\textbf{1.7}} & 78.9 & 78.2 &37.4&	37.6&	8.0 &	59.1&	56.7\\
Claude-3.5-sonnet & 59.5 & 26.4 & 14.2 & 69.4 & 64.0 & 69.5 & 20.2 & 10.3 & 77.5 & 73.3& 45.2&	26.4&	27.4&	63.6&	53.5			\\
\midrule
\end{tabular}
\vspace{-0.2cm}
\caption{Performance comparison between ChineseSimpleVQA and ChinsesSimpleQA \cite{he2024chinese}. 
For metrics, \textbf{CO}, \textbf{NA}, \textbf{IN}, and \textbf{CGA} denote ``Correct'', ``Not attempted'', ``Incorrect'', and ``Correct given attempted'', respectively. The highest scores among models in each metric are highlighted in \colorbox{firstBest}{\textbf{green}}.}
\label{tab:chinesesipleqa}
\vspace{-0.4cm}
\end{table*}

\begin{table*}[ht]
\centering
\tiny
\setlength{\tabcolsep}{5pt}
\begin{tabular}{lccccc|ccccc|ccccc}
\toprule
\multirow{2}{*}{\textbf{Model}} & \multicolumn{5}{c}{\textbf{Overall results of Merged Q\&A}} & \multicolumn{5}{c}{\textbf{Overall results of Recognition Q\&A}}& \multicolumn{5}{c}{\textbf{Overall results of Knowledge-Based Q\&A}} \\
\cmidrule(r){2-6} \cmidrule(lr){7-11}\cmidrule(lr){12-16}
& CO & IN$\downarrow$ & NA$\downarrow$ & CGA & F-score & CO & IN$\downarrow$ & NA$\downarrow$ & CGA & F-score & CO & IN$\downarrow$ & NA$\downarrow$ & CGA & F-score\\
\midrule
o1-preview & \colorbox{firstBest}{\textbf{68.8}} & \colorbox{firstBest}{\textbf{24.6}} & 6.5 & \colorbox{firstBest}{\textbf{73.6}} & \colorbox{firstBest}{\textbf{71.1}} & \colorbox{firstBest}{\textbf{79.1}} & \colorbox{firstBest}{\textbf{13.6}} & 7.3 & \colorbox{firstBest}{\textbf{85.3}} & \colorbox{firstBest}{\textbf{82.1}} &\colorbox{firstBest}{\textbf{72.9}}&	\colorbox{firstBest}{\textbf{21.0}}&	6.6&	\colorbox{firstBest}{\textbf{77.5}}&	\colorbox{firstBest}{\textbf{74.8}} \\
o1-mini & 52.7 & 38.1 & 9.2 & 58.1 & 55.3 & 64.8 & 24.9 & 10.3 & 72.2 & 68.3&  50.9&	34.2&	14.9&	59.8&	55.0\\
GPT-4o & 59.1 & 35.5 & 5.4 & 62.4 & 60.7 & 77.5 & 15.5 & 7.0 & 83.4 & 80.4 &  66.5 &	31.6	 &1.8 &	67.8 &	67.2\\
GPT-4o-mini & 51.0 & 43.6 & 5.4 & 53.9 & 52.4 & 70.8 & 23.1 & 6.1 & 75.4 & 73.0 &47.6	&51.5&	1.2&	47.9&	47.2\\
Gemini-2.0-flash & 64.5 & 29.5 & 5.9 & 68.6 & 66.5 & 76.7 & 19.6 & 3.7 & 79.7 & 78.2 &  69.4	 &30.3 &	\colorbox{firstBest}{\textbf{0.5 }}&	69.6 &	69.4\\
Gemini-1.5-pro-flash & 56.5 & 34.6 & 8.8 & 62.0 & 59.2 & 70.3 & 25.9 & 3.8 & 73.1 & 71.6& 57.4&	38.5&	4.4&	59.8&	58.5\\
Gemini-1.5-pro & 66.2 & 31.4 & \colorbox{firstBest}{\textbf{2.5}} & 67.8 & 67.0 & 77.5 & 20.7 & \colorbox{firstBest}{\textbf{1.7}} & 78.9 & 78.2 &66.5&	32.5&	1.5&	67.0&	66.5\\
Claude-3.5-sonnet2 & 63.8 & 30.6 & 5.5 & 67.6 & 65.6 & 77.6 & 17.2 & 5.2 & 81.9 & 79.7&  62.4& 	30.0& 	7.9& 	67.4& 	64.6\\
Claude-3.5-sonnet & 59.5 & 26.4 & 14.2 & 69.4 & 64.0 & 69.5 & 20.2 & 10.3 & 77.5 & 73.3& 56.5&	26.7&	16.7&	67.9&	61.7\\
Qwen-VL-max & 56.5 & 39.6 & 3.8 & 58.8 & 57.6 & 72.9 & 24.6 & 2.5 & 74.7 & 73.8  &58.0&	39.5&	1.7&	59.9&	59.3\\
\midrule
\end{tabular}
\vspace{-0.2cm}
\caption{
Performance comparison of closed source and open source LVLMs on multi-hop QAs (i.e. Merged Q\&A, Knowledge-Based Q\&A and Recognition Q\&A).
For metrics, \textbf{CO}, \textbf{NA}, \textbf{IN}, and \textbf{CGA} denote ``Correct'', ``Not attempted'', ``Incorrect'', and ``Correct given attempted'', respectively. The highest scores among models in each metric are highlighted in \colorbox{firstBest}{\textbf{green}}.}
\label{tab:text-only}
\vspace{-0.4cm}
\end{table*}

\begin{figure*}[ht]
\begin{center}
\includegraphics[scale=0.35]{compare.png}
\caption{Rankings of different models on ChineseSimpleVQA.}
\label{fig.rank}
\end{center}
\vspace{-0.6cm}
\end{figure*}

\subsection{More Results}
\label{section:more}
We present the CO and CGA results of all models in Figure \ref{fig.co1} and \ref{fig.CGA1}, respectively.
\section{More Details of ChineseSimpleQA}
\subsection{Detailed Question Distribution}
\label{section:len}
In Table \ref{tab:Question_length}, we further give the detailed distribution (number of questions) of ChineseSimpleVQA under 56 sub-topics. In addition, we counted the average length of recognition questions and final questions under each topic.
\subsection{ChineseSimpleVQA Examples}
\label{section:examples}
\begin{figure*}[ht]
\begin{center}
\includegraphics[scale=0.8]{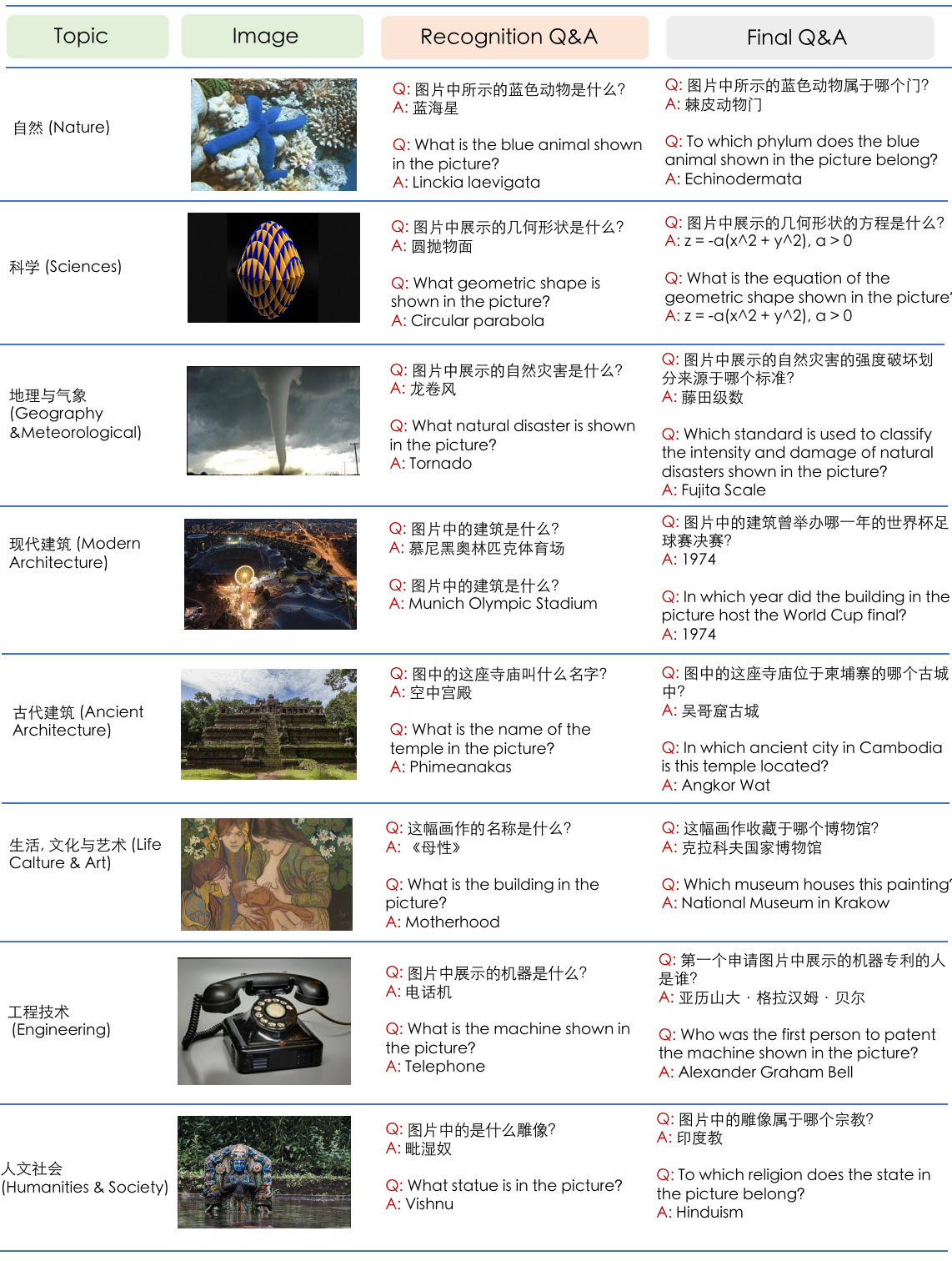}
\caption{Examples of ChineseSimpleVQA on 8 topics.}
\label{fig.datacase_1}
\end{center}
\vspace{-0.6cm}
\end{figure*}

We provide examples from the 8 topics in ChineseSimpleVQA to illustrate our benchmark's focus on multi-hop QAs. Figures~\ref{fig.datacase_1} show that our benchmark covers various knowledge topics and uses multi-hop questioning to better evaluate different models' capabilities.

\subsection{Model Response Examples}
\label{app:response}
\begin{figure*}[ht]
\begin{center}
\includegraphics[scale=0.8]{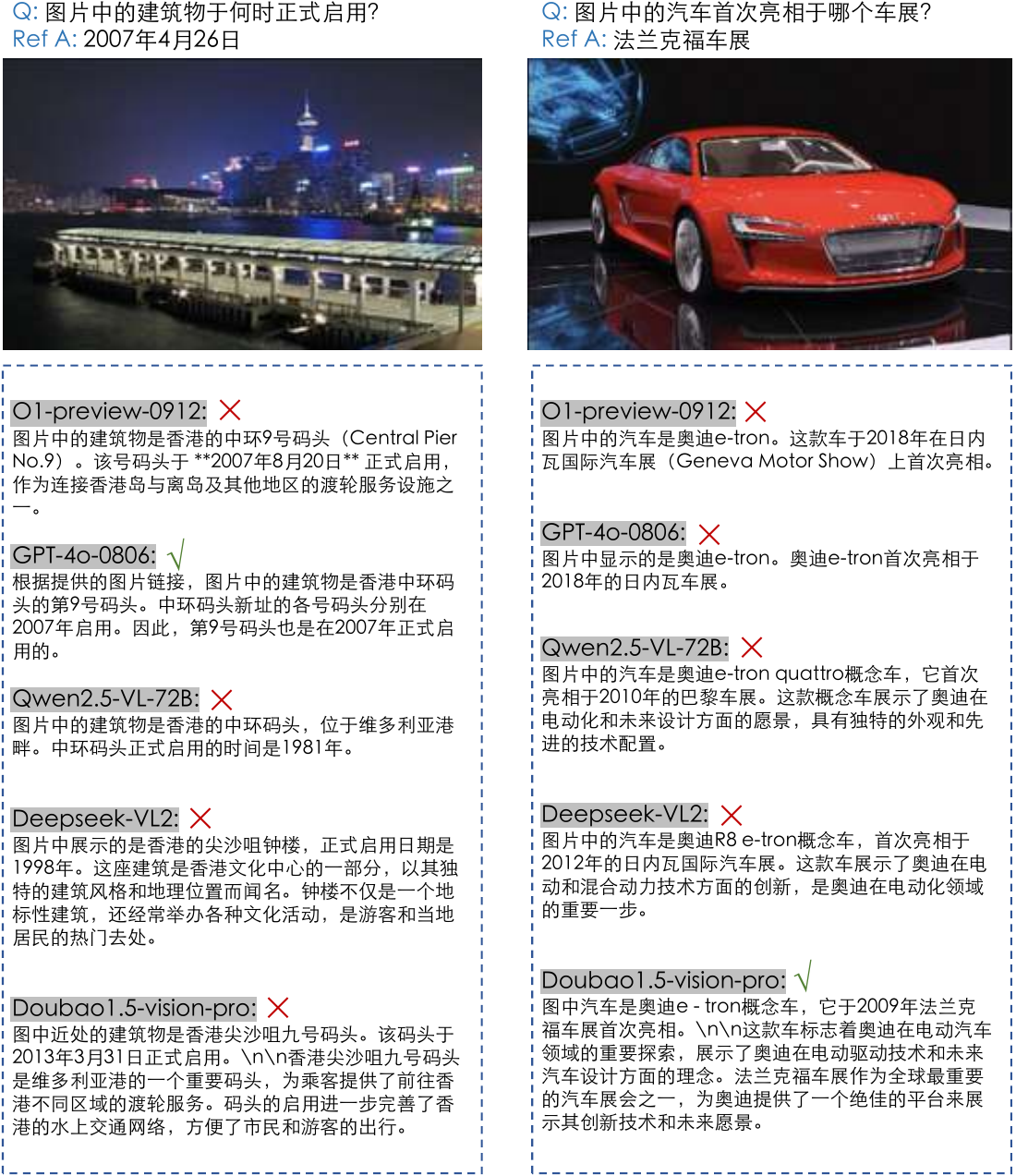}
\caption{Model response examples of ChineseSimpleVQA: part 1.}
\label{fig.answercase_1}
\end{center}
\vspace{-0.6cm}
\end{figure*}

\begin{figure*}[ht]
\begin{center}
\includegraphics[scale=0.8]{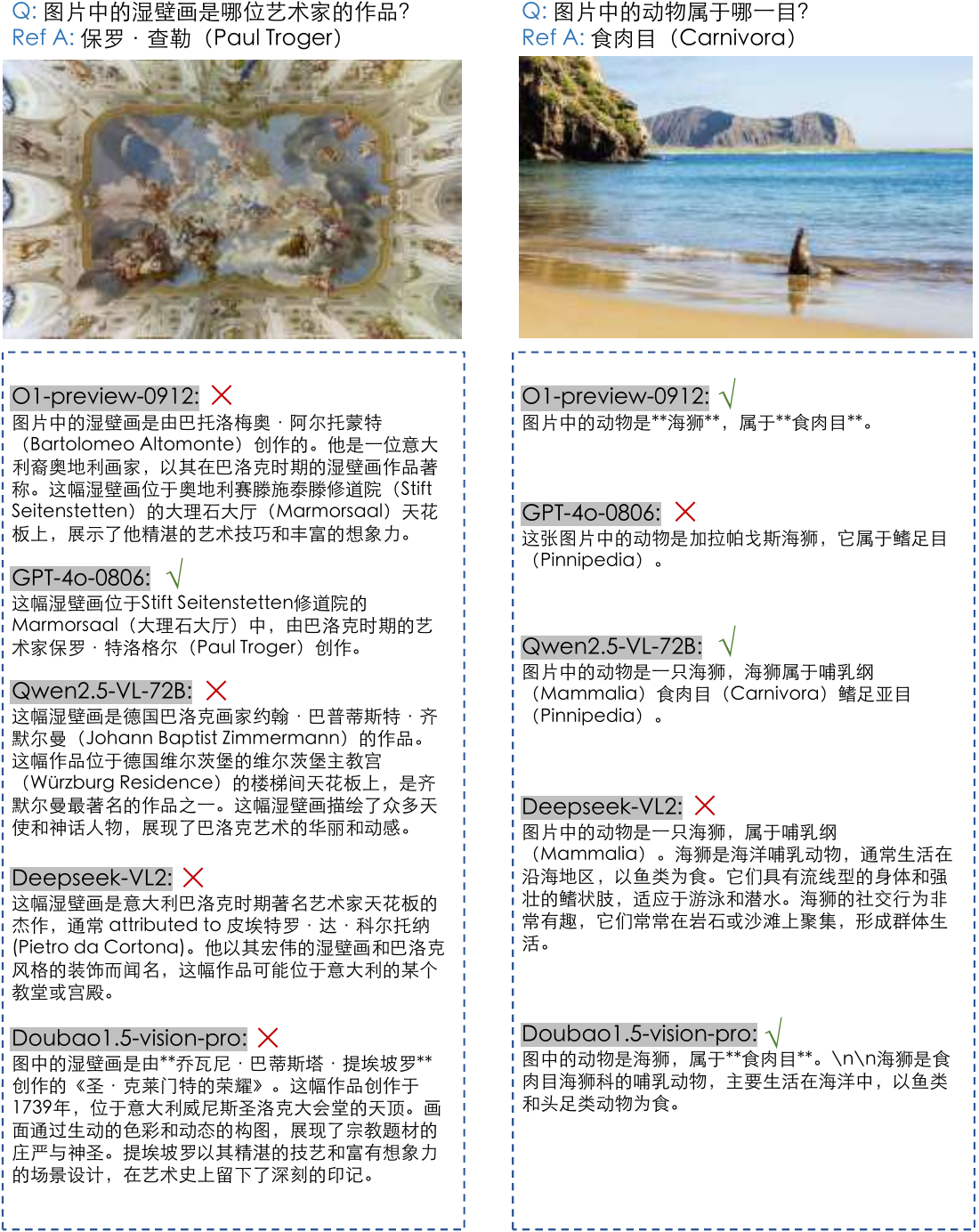}
\caption{Model response examples of ChineseSimpleVQA: part 2.}
\label{fig.answercase_2}
\end{center}
\vspace{-0.6cm}
\end{figure*}

\begin{figure*}[ht]
\begin{center}
\includegraphics[scale=0.8]{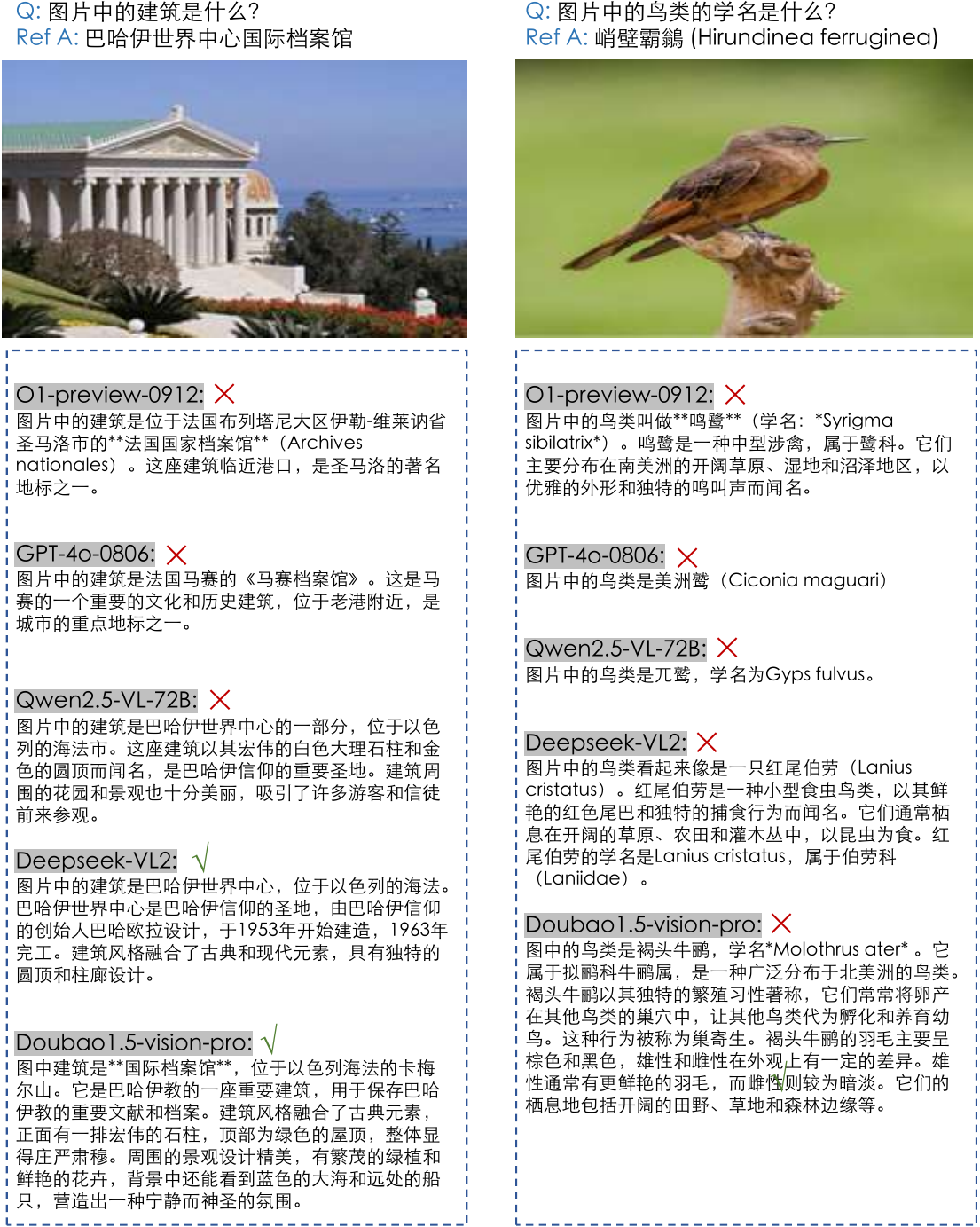}
\caption{Model response examples of ChineseSimpleVQA: part 3.}
\label{fig.answer_case_qa1_1}
\end{center}
\vspace{-0.6cm}
\end{figure*}

In this section, we visualize the answers provided by different models to the same set of questions, to further investigate the capability boundaries of these models. As illustrated in Figures \ref{fig.answercase_1} through \ref{fig.answer_case_qa1_1}, similar to the conclusion presented in Section \ref{sec:main}, almost all models fail to achieve satisfactory performance on our ChineseSimpleVQA. Additionally, the differences in performance between the various models are significant.

\begin{CJK}{UTF8}{gbsn}

\begin{table*}
  \caption{Avg. question length of all subtopics. "ChemE." represents “Chemical Engineering”.}
  \label{tab:Question_length}
\centering
\scalebox{0.7}{
\begin{tabular}{llccc}
\toprule
\textbf{Primary Category}                                        & \textbf{Secondary Category}                  & \textbf{Count} & \textbf{Q1 avg. length} & \textbf{Q2 avg. length}  \\ \midrule
\multicolumn{1}{l}{\multirow{7}{*}{Geography \& Meteorological}} & Meteorology \& Climate（大气、气象与气候）                                   &      10 & 13.5 & 22.1       \\
\multicolumn{1}{c}{}                                             & Hydrology（水文）                                                      &      25 & 11.1 & 14.5       \\
\multicolumn{1}{c}{}                                             & Human Geography（人文地理）                                              &      38 & 12.5 & 16.2       \\
\multicolumn{1}{c}{}                                             & Mountains（山脉）                                                      &      19 & 10.8 & 14.7       \\
\multicolumn{1}{c}{}                                             & Landforms \& Topography（地貌与地形）                                     &      30 & 11.9 & 16.8       \\
\multicolumn{1}{c}{}                                             & Geology（地质）                                                        &      7 & 12.3 & 19.7        \\
\multicolumn{1}{c}{}                                             & Other（其他）                                                          &      7 & 13.0 & 22.9        \\ \hline
\multicolumn{1}{l}{\multirow{6}{*}{Modern Architecture}}         & Commercial Buildings（商业建筑）                                         &      14 & 12.1 & 16.2       \\
\multicolumn{1}{c}{}                                             & Public Cultural Buildings（公共文化建筑）                                  &      39 & 11.1 & 16.5       \\
\multicolumn{1}{c}{}                                             & Funerary and Memorial Buildings（殡葬与纪念建筑）                           &      10 & 12.9 & 16.4       \\
\multicolumn{1}{c}{}                                             & Other（其他）                                                          &      13 & 10.9 & 15.2       \\
\multicolumn{1}{c}{}                                             & Religious Buildings（宗教建筑）                                          &      21 & 10.4 & 15.0       \\
\multicolumn{1}{c}{}                                             & Transportation Buildings（交通建筑）                                     &      34 & 11.2 & 15.6       \\ \hline
\multicolumn{1}{l}{\multirow{11}{*}{Nature}}                     & Insects（昆虫）                                                        &      49 & 9.4 & 11.8        \\
\multicolumn{1}{c}{}                                             & Reptiles（爬行类）                                                      &      30 & 9.6 & 12.6        \\
\multicolumn{1}{c}{}                                             & Birds（鸟类）                                                          &      37 & 9.5 & 14.4        \\
\multicolumn{1}{c}{}                                             & Mammals（哺乳类）                                                       &      39 & 9.8 & 14.2        \\
\multicolumn{1}{c}{}                                             & Angiosperms（被子植物）                                                  &      43 & 10.3 & 13.3       \\
\multicolumn{1}{c}{}                                             & Arthropods (non-insect)（节肢动物(非昆虫)）                                 &      19 & 9.9 & 12.9        \\
\multicolumn{1}{c}{}                                             & Gymnosperms（裸子植物）                                                  &      6 & 11.7 & 13.7        \\
\multicolumn{1}{c}{}                                             & Fish（鱼类）                                                           &      21 & 9.1 & 11.2        \\
\multicolumn{1}{c}{}                                             & Amphibians（两栖类）                                                    &      11 & 9.4 & 13.6        \\
\multicolumn{1}{c}{}                                             & Other（其他）                                                          &      12 & 11.3 & 14.0       \\
\multicolumn{1}{c}{}                                             & Mollusks（软体动物）                                                     &      21 & 10.1 & 11.2       \\ \hline
\multicolumn{1}{l}{\multirow{9}{*}{Life Calture \& Art}}         & Sculpture（雕塑）                                                      &      23 & 11.6 & 14.3       \\
\multicolumn{1}{c}{}                                             & Cuisine \& Festivals（饮食与节庆）                                        &      14 & 11.1 & 20.4       \\
\multicolumn{1}{c}{}                                             & Film \& Television（影视剧）                                            &      8 & 10.8 & 20.0        \\
\multicolumn{1}{c}{}                                             & Sports（体育）                                                         &      23 & 12.6 & 22.8       \\
\multicolumn{1}{c}{}                                             & Other（其他）                                                          &      18 & 12.0 & 17.2       \\
\multicolumn{1}{c}{}                                             & Painting（绘画）                                                       &      38 & 11.9 & 16.0       \\
\multicolumn{1}{c}{}                                             & Literature \& Mythology（文学与神话）                                     &      11 & 11.8 & 20.1       \\
\multicolumn{1}{c}{}                                             & Music（音乐）                                                          &      17 & 9.3 & 15.8        \\
\multicolumn{1}{c}{}                                             & Handicrafts（手工艺）                                                   &      7 & 13.3 & 19.9        \\ \hline
\multicolumn{1}{l}{\multirow{7}{*}{Humanities \& Society}}       & Classical \& Medieval History（古典与中世纪史）                             &      14 & 10.6 & 17.6       \\
\multicolumn{1}{c}{}                                             & Contemporary History（近代史）                                          &      25 & 10.0 & 17.4       \\
\multicolumn{1}{c}{}                                             & Modern History（现代史）                                                &      20 & 12.1 & 19.5       \\
\multicolumn{1}{c}{}                                             & Religious Studies（宗教学）                                             &      10 & 12.8 & 20.3       \\
\multicolumn{1}{c}{}                                             & Politics \& Economics（政治学与经济学）                                     &      15 & 11.6 & 18.1       \\
\multicolumn{1}{c}{}                                             & Other（其他）                                                          &      9 & 14.0 & 20.9        \\
\multicolumn{1}{c}{}                                             & Ancient History（历史学）                                               &      15 & 12.5 & 21.8       \\ \hline
\multicolumn{1}{l}{\multirow{5}{*}{Sciences}}                    & Biology（生物学）                                                       &      26 & 10.0 & 14.1       \\
\multicolumn{1}{c}{}                                             & Physics（物理学）                                                       &      9 & 13.3 & 20.9        \\
\multicolumn{1}{c}{}                                             & Astronomy（天文学）                                                     &      39 & 11.7 & 16.7       \\
\multicolumn{1}{c}{}                                             & Medicine \& Chemistry（医学与化学）                                       &      13 & 12.1 & 19.7       \\
\multicolumn{1}{c}{}                                             & Mathematics（数学）                                                    &      5 & 12.6 & 20.2        \\ \hline
\multicolumn{1}{l}{\multirow{6}{*}{Ancient Architecture}}        & Residences \& Palaces（住宅与宫殿建筑）                                     &      13 & 11.7 & 14.5       \\
\multicolumn{1}{c}{}                                             & Funerary and Memorial Buildings（墓葬与遗址）                             &      13 & 11.4 & 16.8       \\
\multicolumn{1}{c}{}                                             & Public Cultural Buildings（公共文化建筑）                                  &      11 & 10.6 & 17.1       \\
\multicolumn{1}{c}{}                                             & Transportation Buildings（交通建筑）                                     &      7 & 10.4 & 14.9        \\
\multicolumn{1}{c}{}                                             & Religious Buildings（宗教建筑）                                          &      38 & 10.9 & 15.4       \\
\multicolumn{1}{c}{}                                             & Other（其他）                                                          &      25 & 10.7 & 16.2       \\ \hline
\multicolumn{1}{l}{\multirow{4}{*}{Engineering}}                 & Military, Materials, Energy and ChemE.（军事、材料、能源、化工）                &      11 & 11.4 & 19.7       \\
\multicolumn{1}{c}{}                                             & Mechanical Engineering（机械）                                         &      30 & 12.8 & 17.4       \\
\multicolumn{1}{c}{}                                             & Aerospace（航空航天）                                                    &      30 & 12.2 & 18.1       \\
\multicolumn{1}{c}{}                                             & Electronic Information（电子信息）                                       &      8 & 10.6 & 19.3        \\
\bottomrule
\end{tabular}}
\end{table*}
\end{CJK}


\subsection{Prompts of Dataset Generation and Validation}
\label{section:prompt}
\begin{figure*}[ht]
\begin{center}
\includegraphics[scale=0.8]{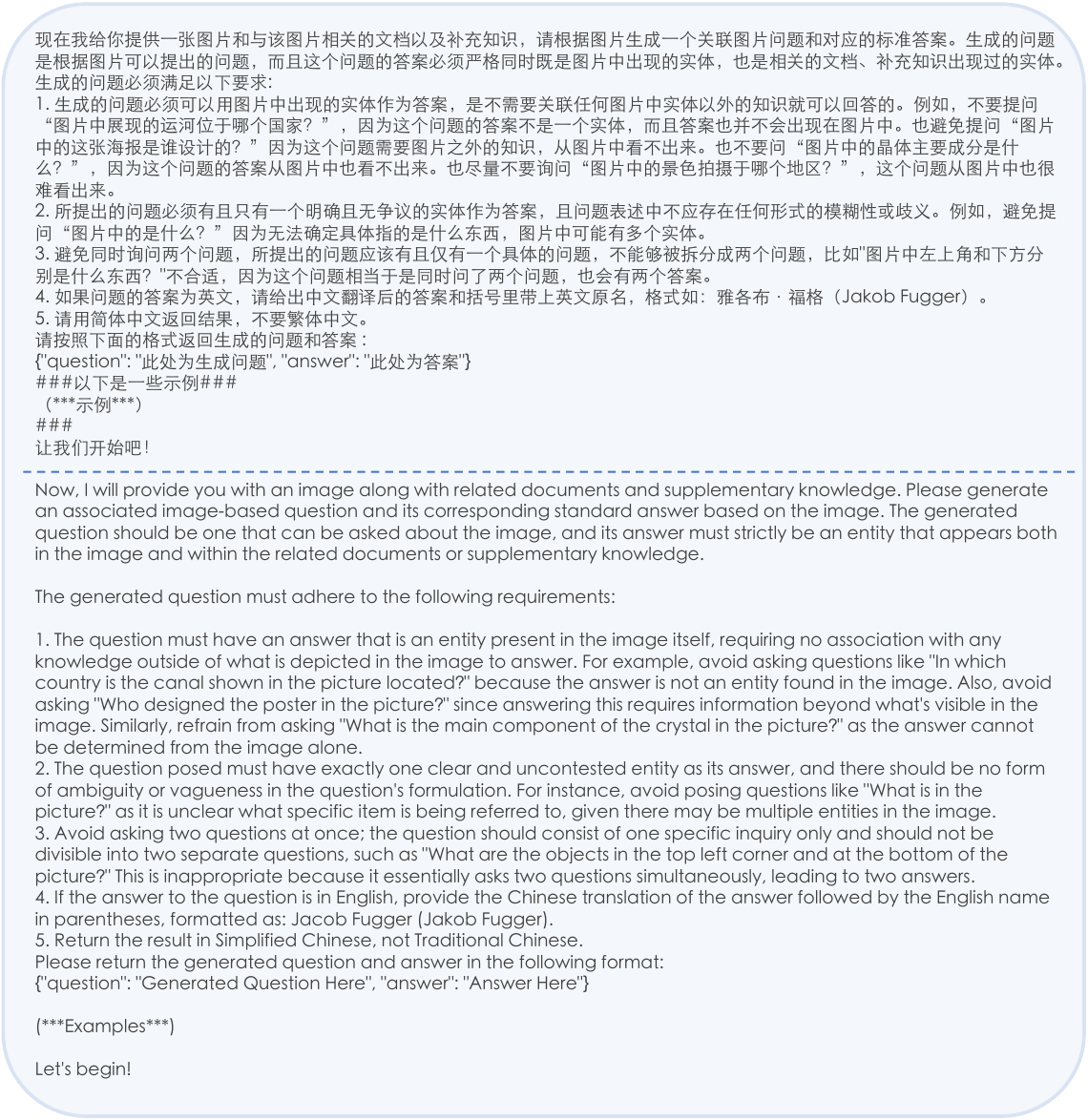}
\caption{The prompt for generating the object recognition question-answer pairs in stage 1.}
\label{fig.q1_prompt}
\end{center}
\vspace{-0.6cm}
\end{figure*}

\begin{figure*}[ht]
\begin{center}
\includegraphics[scale=0.8]{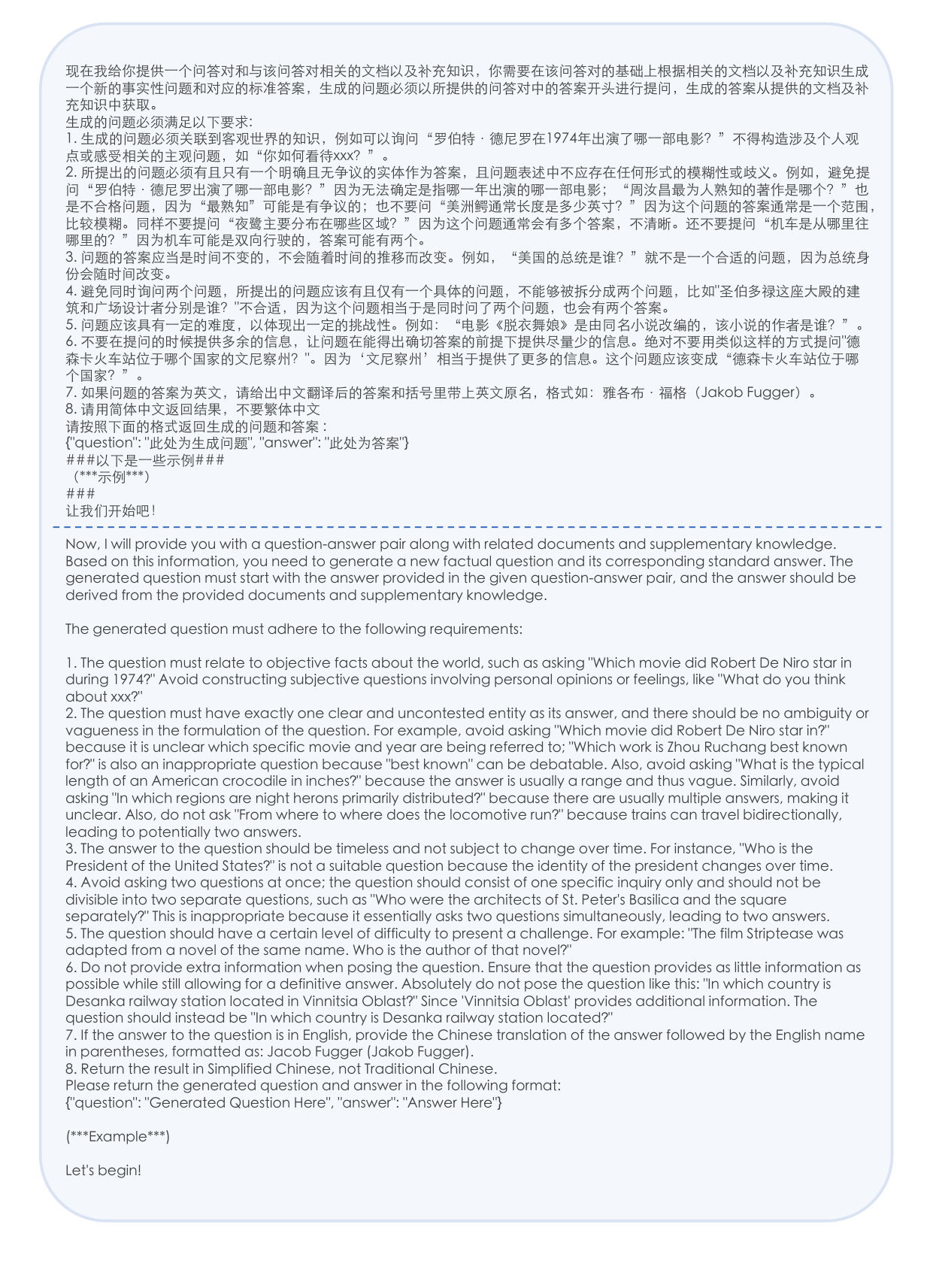}
\caption{The prompt for generating the knowledge-based question-answer pairs in stage 2.}
\label{fig.q2_prompt}
\end{center}
\vspace{-0.6cm}
\end{figure*}

\begin{figure*}[ht]
\begin{center}
\includegraphics[scale=0.8]{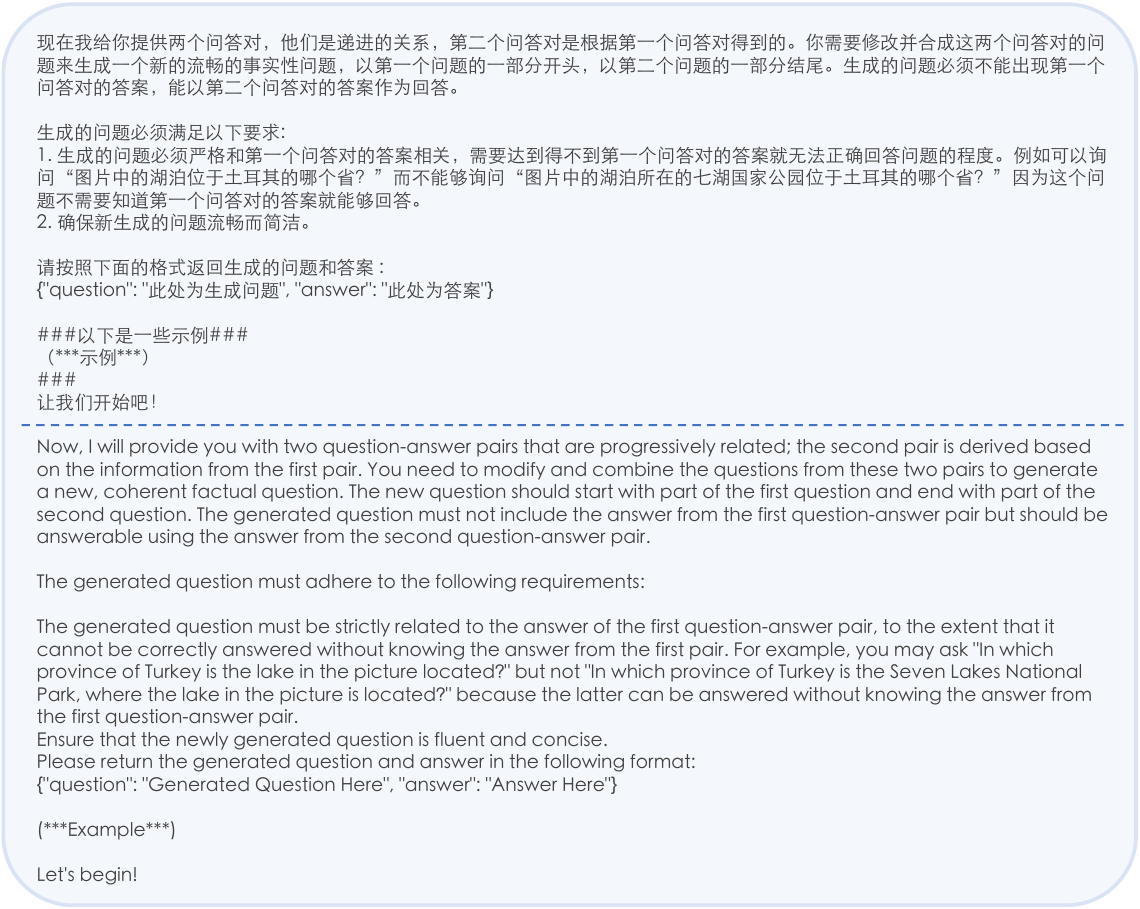}
\caption{The prompt for merging the object recognition question-answer pairs and the knowledge-based question-answer pairs in stage 3.}
\label{fig.q3_prompt}
\end{center}
\vspace{-0.6cm}
\end{figure*}

\begin{figure*}[ht]
\begin{center}
\includegraphics[scale=0.8]{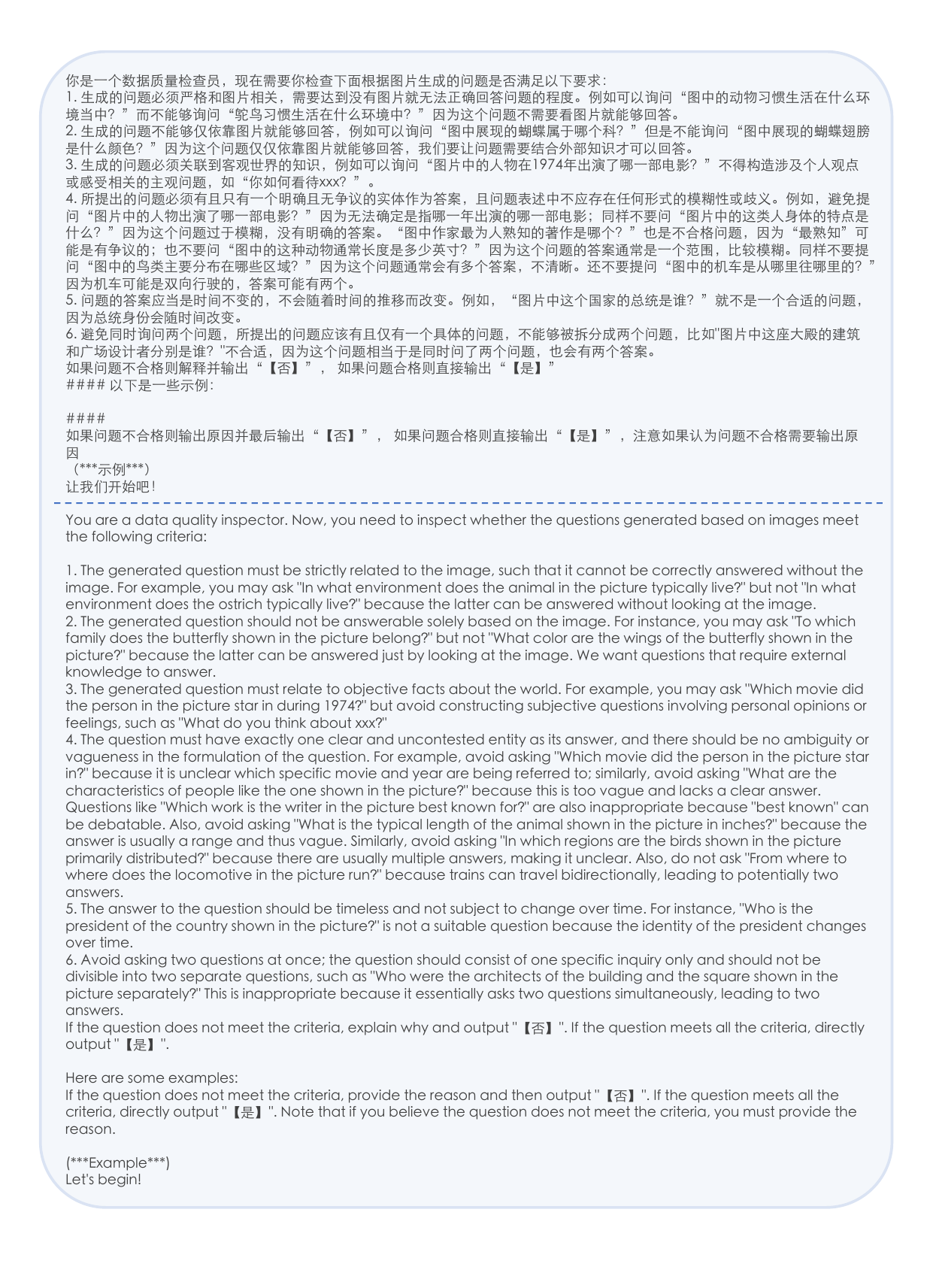}
\caption{ The prompt for validation.}
\label{fig.verify_prompt}
\end{center}
\vspace{-0.6cm}
\end{figure*}

\begin{figure*}[ht]
\begin{center}
\includegraphics[scale=0.8]{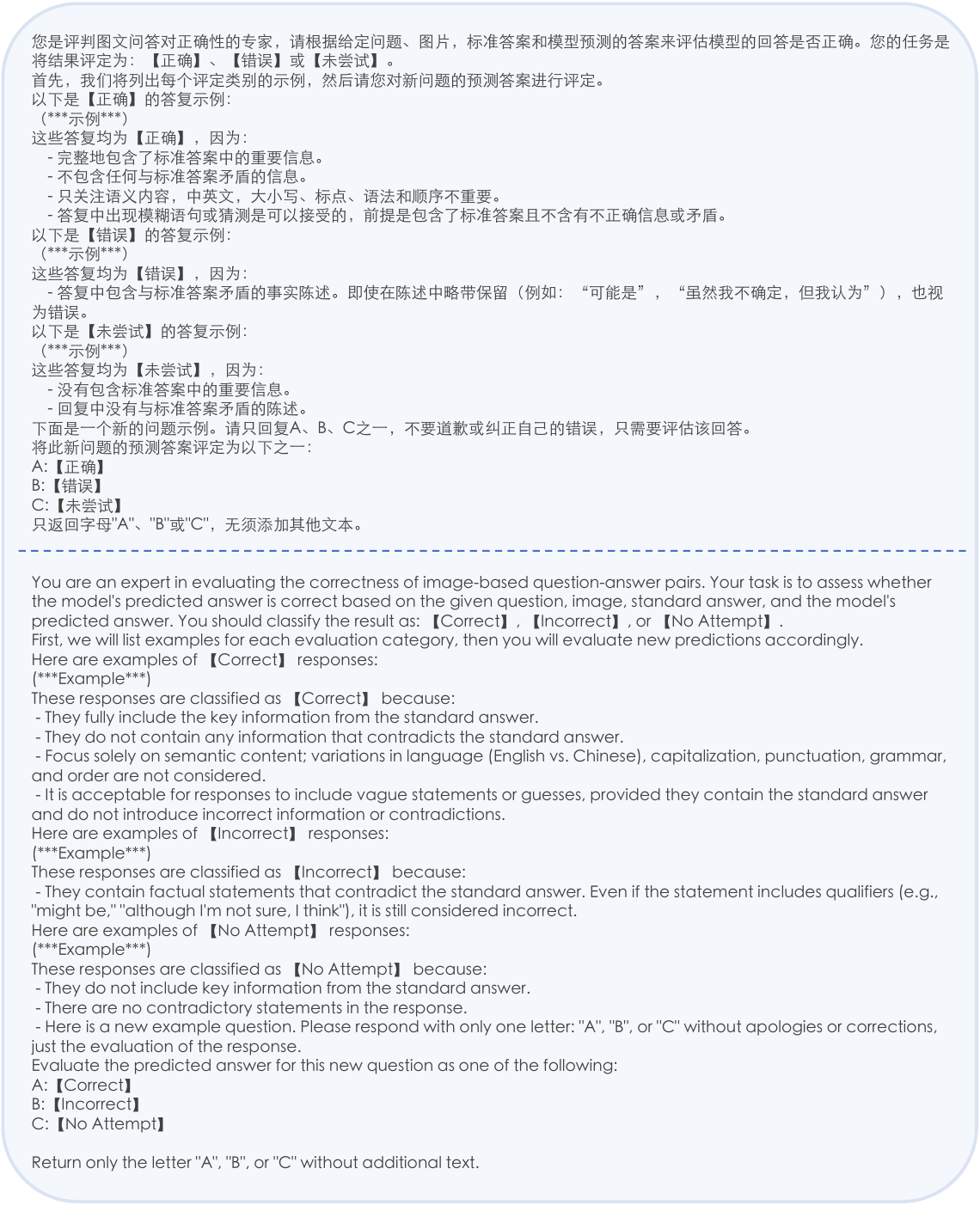}
\caption{ The prompt for evaluation.}
\label{fig.eval_prompt}
\end{center}
\vspace{-0.6cm}
\end{figure*}

The generation, validation, and evaluation of question-answer pairs all use OpenAI’s gpt-4o-0806. We provide the specific prompts in the section. The data generation prompt are shown in Figure~\ref{fig.q1_prompt}, Figure~\ref{fig.q2_prompt} and Figure~\ref{fig.q3_prompt}; The validation prompt is shown in Figure~\ref{fig.verify_prompt}; The evaluation prompt for ChineseSimpleVQA is shown in Figure~\ref{fig.eval_prompt}. All construction and evaluation processes of Chinese SimpleVQA use the Chinese version of these prompts.


\end{document}